\documentclass{article}

\usepackage[final]{neurips_2022}

\usepackage[utf8]{inputenc}
\usepackage[T1]{fontenc}
\usepackage{url}
\usepackage{booktabs}
\usepackage{amsfonts}
\usepackage{nicefrac}
\usepackage{microtype}
\usepackage[hidelinks]{hyperref}
\hypersetup{
    colorlinks = true,
    linkcolor = blue,
    urlcolor  = blue,
    citecolor = blue,
    anchorcolor = blue}
\usepackage{listings}
\usepackage[inline]{enumitem}
\usepackage[table, dvipsnames]{xcolor}
\usepackage{graphicx,wrapfig}
\usepackage{amsmath}
\usepackage{stmaryrd}
\usepackage{float}
\usepackage[export]{adjustbox}
\usepackage{caption}
\usepackage{subcaption}
\usepackage{mathtools, nccmath}
\usepackage{tikz}
\usepackage{multicol}
\usepackage{multirow}
\usepackage{diagbox}
\usepackage{wrapfig}
\usepackage{algorithm}
\usepackage[algo2e, ruled,vlined, boxed, linesnumbered]{algorithm2e}
\SetArgSty{textnormal}
\usepackage[colorinlistoftodos]{todonotes}
\usepackage{setspace}
\usepackage{amsmath,amsfonts,bm}

\def\1{\bm{1}}

\DeclareMathAlphabet{\mathsfit}{\encodingdefault}{\sfdefault}{m}{sl}
\SetMathAlphabet{\mathsfit}{bold}{\encodingdefault}{\sfdefault}{bx}{n}

\newcommand{\R}{\mathbb{R}}

\newcommand{\defeq}{\vcentcolon=}

\def\bP{\mathbf{P}}

\def\RR{\mathbb{R}}

\def\We{W_{\varepsilon}}

\DeclarePairedDelimiterX{\dotp}[2]{\langle}{\rangle}{#1, #2}

\definecolor{darkblue}{HTML}{1A254B}
\definecolor{lightblue}{HTML}{A7BED3}
\definecolor{blue}{HTML}{2B50AA}
\definecolor{green}{HTML}{81B5AE}
\definecolor{pink}{HTML}{F2545B}
\definecolor{red}{HTML}{A4243B}
\definecolor{lightgray}{HTML}{E5EBEF}
\def\bpot{\psi}
\def\cont{c}
\def\RR{\mathbb{R}}
\def\Ccal{\mathcal{C}}

\def\bm{\mathbf{m}}
\newcommand{\newinf}{\mathop{\mathrm{inf}\vphantom{\mathrm{sup}}}}
\lstdefinestyle{codestyle}{
    commentstyle=\color{blue},
    keywordstyle=\color{lightblue},
    numberstyle=\tiny\color{gray},
    stringstyle=\color{pink},
    basicstyle=\ttfamily\footnotesize,
    breakatwhitespace=false,         
    breaklines=true,                 
    captionpos=b,                    
    keepspaces=true,                 
    showspaces=false,                
    showstringspaces=false,
    showtabs=true,                  
    tabsize=2,
    frame=leftline
}
\newcommand{\newtext}[1]{{\color{black} #1}}

\title{Supervised Training of Conditional Monge Maps}

\author{%
  Charlotte Bunne\thanks{Work done during an internship at Apple.} \\
  ETH Zurich\\
  \texttt{bunnec@ethz.ch} \\
  \And
  Andreas Krause \\
  ETH Zurich \\
  \texttt{krausea@ethz.ch} \\
  \And
  Marco Cuturi \\
  Apple \\
  \texttt{cuturi@apple.com} 
}

\begin{document}
\abovedisplayskip=1.5pt
\abovedisplayshortskip=1.5pt
\belowdisplayskip=1.5pt
\belowdisplayshortskip=1.5pt
\abovecaptionskip=5pt
\belowcaptionskip=1pt
\renewcommand\ttdefault{lmtt}
\newcommand*{\tabindent}{\hspace{3mm}}

\maketitle


\begin{abstract}
\looseness -1
Optimal transport (OT) theory describes general principles to define and select, among many possible choices, the most efficient way to map a probability measure onto another.
That theory has been mostly used to estimate, given a pair of source and target probability measures $(\mu,\nu)$, a parameterized map $T_\theta$ that can efficiently map $\mu$ onto $\nu$.
In many applications, such as predicting cell responses to treatments, pairs of input/output data measures $(\mu,\nu)$ that define optimal transport problems do not arise in isolation but are associated with a \textit{context} $c$, as for instance a treatment when comparing populations of untreated and treated cells.
To account for that context in OT estimation, we introduce \textsc{CondOT}, a multi-task approach to estimate a family of OT maps conditioned on a context variable, using several pairs of measures $(\mu_i, \nu_i)$ tagged with a context label~$c_i$. 
\textsc{CondOT} learns a \textit{global} map $\mathcal{T}_{\theta}$ conditioned on context that is not only expected to fit {\em all labeled pairs} in the dataset $\{(c_i, (\mu_i, \nu_i))\}$, i.e., $\mathcal{T}_{\theta}(c_i) \sharp\mu_i \approx \nu_i$, but should also \textit{generalize} to produce meaningful maps $\mathcal{T}_{\theta}(c_{\text{new}})$ when conditioned on unseen contexts $c_{\text{new}}$. Our approach harnesses and provides a novel usage for {\em partially input convex neural networks}, for which we introduce a robust and efficient initialization strategy inspired by Gaussian approximations. We demonstrate the ability of \textsc{CondOT} to infer the effect of an arbitrary combination of genetic or therapeutic perturbations on single cells, using only observations of the effects of said perturbations separately.
\end{abstract}

\section{Introduction}
\looseness -1 A key challenge in the treatment of cancer is to predict the effect of drugs, or a combination thereof, on cells of a particular patient. To achieve that goal, single-cell sequencing can now provide measurements for individual cells, in treated and untreated conditions, but these are, however, not in correspondence.  Given such examples of untreated and treated cells under different drugs, can we predict the effect of new drug combinations?  We develop a general approach motivated by this and related problems, through the lens of {\em optimal transport (OT) theory}, and, in that process, develop tools that might be of interest for other application domains of OT. Given a collection of $N$ pairs of measures $(\mu_i,\nu_i)$ over $\R^d$ (cell measurements), tagged with a context $c_i$ (encoding the treatment), we seek to learn a context-dependent, parameterized transport map $\mathcal{T}_{\theta}$ such that, on training data, that map $\mathcal{T}_{\theta}(c_i):\mathbb{R}^d\rightarrow\mathbb{R}^d$ fits the dataset, in the sense that $\mathcal{T}_{\theta}(c_i) \sharp\mu_i \approx \nu_i$. Additionally, we expect that this parameterized map can generalize to unseen contexts and patients, to predict, given a patient's cells described in $\mu_{\text{new}}$, the effect of applying context $c_{\text{new}}$ on these cells as $\mathcal{T}_{\theta}(c_{\text{new}})\sharp\mu$.

\looseness -1 \textbf{Learning Mappings Between Measures }
From generative adversarial networks, to normalizing flows and diffusion models, the problem of learning maps that move points from a source to a target distribution is central to machine learning. OT theory~\citep{santambrogio2015optimal} has emerged as a principled approach to carry out that task: For a pair of measures $\mu,\nu$ supported on $\mathbb{R}^d$, OT suggests that, among all maps $T$ such that $\nu$ can be reconstructed by applying $T$ to every point in the support of $\mu$ (abbreviated with the push-forward notation as $T\sharp\mu=\nu$), one should favor so-called \citeauthor{Monge1781} maps, which \textit{minimize} the average squared-lengths of displacements $\|x-T(x)\|^{2}$. A rich literature, covered in \citet{Peyre2019computational}, addresses computational challenges of estimating such maps, with impactful applications to various areas of science \citep[cf.,][]{hashimoto2016learning,schmitz2018wasserstein,schiebinger2019,yang2020predicting,janati2020multi,bunne2022recovering}.

\looseness -1 \textbf{Neural OT }  We focus in this work on neural approaches that parameterize the optimal maps $T$ as neural networks. An early approach is the work on Wasserstein GANs~\citep{arjovsky2017wasserstein}, albeit the transport map is not explicitly estimated. 
Several recent results have exploited a more explicit connection between OT and NNs, derived from the celebrated \citeauthor{Brenier1987} theorem \citeyearpar{Brenier1987}, which states that Monge maps are necessarily gradients of convex functions.
Such convex functions can be represented using input convex neural networks (ICNN)~\citep{amos2017input}, to parameterize either the Monge map~\citep{jacob2018w2gan,yang2018scalable,bunne2021learning,bunne2022proximal} or a dual potential~\citep{makkuva2020optimal,korotin2020wasserstein} as, respectively, the gradient of an ICNN or an ICNN itself.
In this paper, we build on this line of work, but substantially generalize it, to learn a {\em parametric} family of context-aware transport maps, using a collection of labeled pairs of measures.

\begin{figure*}
    \centering
    \includegraphics[width= \textwidth]{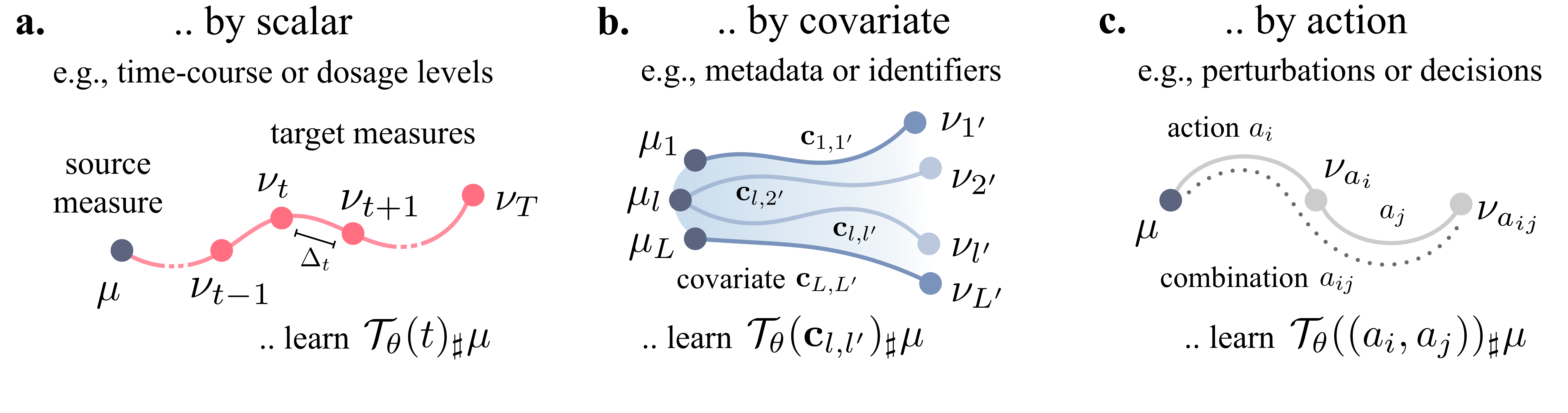}
    \caption{The evolution from a source $\mu$ to a target measure $\nu$ can depend on context variables $c$ of various nature. This comprises \textbf{a.} scalars such as time or dosage $t$ which determine the magnitude of an optimal transport, \textbf{b.} flow of measures into another one based on additional information \newtext{(possibly different between $\mu$ and $\nu$) stored in vectors $\mathbf{c}_{l,l^\prime}$}, or \textbf{c.} discrete and complex actions $a_i$, possibly in combination $a_{ij}$. We seek a unified framework to produce a map $\mathcal{T}_\theta(c)$ from any type of condition $c$. \vspace{-15pt}}
    \label{fig:overview}
\end{figure*}

\looseness -1 \textbf{Contributions }
We propose a framework that can leverage \textit{labeled} pairs of measures $\{(\cont_i, (\mu_i, \nu_i))\}_i$ to infer a \textit{global} parameterized map $\mathcal{T}_{\theta}$. 
Hereby, the context $\cont_i$ belongs to an arbitrary set $\mathcal{C}$. We construct $\mathcal{T}_{\theta}$ so that it should be able, given a possibly unseen context label $\cont\in\Ccal$, to output a map $\mathcal{T}_{\theta}(c):\mathbb{R}^d\rightarrow\mathbb{R}^d$, that is itself the gradient of a convex function. To that end, we propose to learn these parameterized Monge maps $\mathcal{T}_{\theta}$ as the gradients of partially input convex neural networks (PICNN), which we borrow from the foundational work of \citet{amos2017input}. 
Our framework can be also interpreted as a hypernetwork \citep{ha2016hypernetworks}: The PICNN architecture can be seen as an ICNN whose weights and biases are \textit{modulated} by the context vector $\cont$, which parameterizes a \textit{family} of convex potentials in $\mathbb{R}^d$.
Because both ICNN ---and to a greater extent PICNN--- are notoriously difficult to train~\citep{richter2021input,korotin2020wasserstein,korotin2021neural}, we use closed-form solutions between Gaussian approximations to derive relevant parameter initializations for (P)ICNNs:
These choices ensure that, \textit{upon initialization}, the gradient of the (P)ICNNs mimics the affine Monge map obtained in closed form between Gaussian approximations of measures $\mu_i,\nu_i$~\citep{gelbrich1990formula}.
Our framework is applied to three scenarios: Parameterization of transport through a real variable (time or drug dosage), through an auxiliary informative variable (cell covariates) and through action variables (genetic perturbations in combination) (see Fig.~\ref{fig:overview}). Our results demonstrate the ability of our architectures to better capture on out-of-sample observations the effects of these variables in various settings, even when considering never-seen, composite context labels. These results suggest potential applications of conditional OT to model personalized medicine outcomes, or to guide novel experiments, where OT could serve as a predictor for never tested context labels.

\section{Background on Neural Solvers for the 2-Wasserstein Problem} \label{sec:background}

\paragraph{Optimal Transport}
The Monge problem between two measures $\mu, \nu\in \mathcal{P}(\mathbb{R}^d)$, here restricted to measures supported on $\mathbb{R}^d$ and compared with the squared Euclidean metric, reads
\begin{equation}\label{eq:monge}
T^\star := \arg\inf_{T\sharp\mu=\nu}\int_{\mathbb{R}^d}\|x-T(x)\|^2d\mu(x)\,.
\end{equation}
The existence of $T^\star$ is guaranteed under fairly general conditions\newtext{~\citep[Theorem 1.22]{santambrogio2015optimal}, which require that $\mu$ and $\nu$ have finite $L_2$ norm, and that $\mu$ puts no mass on $(d-1)$ surfaces of class $\mathcal{C}_2$}. This can be proved with the celebrated \citeauthor{Brenier1987} theorem \citeyearpar{Brenier1987}, which states that there must exist  a unique (up to the addition of a constant) potential $\newtext{f^\star}:\mathbb{R}^d\rightarrow \mathbb{R}$ such that $T^\star = \nabla f^\star$. 
This theorem has far-reaching implications: It is sufficient, when seeking optimal transport maps, to restrict the computational effort to seek a ``good'' convex potential, such that its gradient pushes $\mu$ towards $\nu$. This result has been exploited to propose OT solvers that rely on input convex neural networks (ICNNs)~\citep{amos2017input}, introduced below
\begin{equation}\label{eq:dual}
f^\star:=\arg\!\!\sup_{f\,\text{convex}}\mathcal{E}_{\mu,\nu}(f):=\int_{\mathbb{R}^d}f^*\textrm{d}\mu+\int_{\mathbb{R}^d}f\textrm{d}\nu.
\end{equation}
In practice, Monge maps can be estimated using a dual formulation~\citep{makkuva2020optimal, korotin2020wasserstein, bunne2022proximal, alvarez2021optimizing, mokrov2021large}. 
Indeed, $T^\star$ in~\eqref{eq:monge} is recovered as $\nabla f^\star$, where $f^\star$ is defined in~\eqref{eq:dual}, writing $f^*$ for the Legendre transform of $f$.

\paragraph{Convex Neural Architectures} \label{sec:icnn}
Input convex neural networks (ICNN) are neural networks $\psi_\theta$ that admit certain constraints on their architecture and parameters $\theta$, such that their output $\psi_\theta(x)$ is a convex function of their input $x$~\citep{amos2017input}. As a result, they have been increasingly used as drop-in replacements to the set of admissible functions in~\eqref{eq:dual}. Practically speaking, an ICNN is a $K$-layer, fully connected network such that, at each layer index $k$ from $0$ to $K-1$, a hidden state vector $z_k$ is defined recursively as in~\eqref{eq:icnn},
\begin{equation} \label{eq:icnn}
    z_{k+1} = \sigma_k(W^x_k x + W^z_k z_k + b_k)
\end{equation}
 and $\psi_\theta(x) = z_K$,
where, by convention, $z_0$ and $W^z_0$ are $0$; $\sigma_k$ are \textit{convex} non-decreasing activation functions; $\theta=\{b_k, W^z_k, W^x_k\}_{k=0}^{K-1}$ are the weights and biases of the neural network. While ample flexibility is provided to choose dimensions for intermediate hidden states $z_k$, the last layer must necessarily produce a scalar, hence $W^x_{K-1}$ and $W^z_{k-1}$ are line vectors and $b_{K-1}\in\mathbb{R}$. ICNNs are characterized by the fact that all weight matrices $W^z_k$ associated to latent representations $z$ must have \textit{non-negative} entries. This, along with the specific activation functions, ensures the convexity of $\psi_\theta$. We encode this constraint by identifying these matrices as the elementwise softplus or ReLU of other matrices of the same size, or, alternatively, using a regularizer that penalizes the negative entries of these matrices. Since the work by \citet{amos2017input}, convex neural architectures have been used within the context of OT to model convex dual functions~\citep{makkuva2020optimal}, or normalizing flows derived from convex potentials~\citep{huang2021convex}. Their expressivity and universal approximation properties have been studied by~\citet{chen2018optimal}, who show that any convex function over a compact convex domain can be approximated in sup norm by an ICNN.

\section{Supervised Training of Conditional Monge Maps} \label{sec:method}
We are given a dataset of $N$ pairs of measures, each endowed with a label, $(c_i,(\mu_i, \nu_i))\in\mathcal{C}\times \mathcal{P}(\RR^d)^2$. Our framework builds upon two pillars: (i.) we formulate the hypothesis that an optimal transport $T^\star_i$ (or, equivalently, the gradient of a convex potential $f^\star_i$) explains how measure $\mu_i$ was mapped to $\nu_i$, given context $c_i$; (ii.) we build on the multi-task hypothesis \citep{caruana1997multitask} that all of the $N$ maps $T^\star_i$ between $\mu_i$ and $\nu_i$ share a common set of parameters, that are \textit{modulated} by context informations $c_i$. These ideas are summarized in an abstract regression model described below.

\subsection{A Regression Formulation for Conditional OT Estimation}
\looseness -1 $\theta\in\Theta\subset\mathbb{R}^r$, $\mathcal{T}_\theta$ describes a function that takes an input vector $c\in\mathcal{C}$, and outputs a \textit{function} $\mathcal{T}_\theta(c):\mathbb{R}^d\rightarrow\mathbb{R}^d$, as a hypernetwork would~\citep{ha2016hypernetworks}. Assume momentarily that we are given \textit{ground truth} maps $T_i$, that describe the effect of context $c_i$ on any measure, rather only pairs of measures $(\mu_i,\nu_i)$. This is of course a major leap of faith, since even recovering an OT map $T^\star$ from two measures is in itself very challenging~\citep{hutter2021minimax,rigollet2022sample,pooladian2021entropic}. If such maps were available, a direct supervised approach to learn a unique $\theta$ could hypothetically involve minimizing a fit function composed of losses between maps
\begin{align}\label{eq:newobj1}
\min_\theta \sum_{i=1}^N \int_{\mathbb{R}^d} \|\mathcal{T}_{\theta}(c_i)(x) - T_i(x)\|^2\, \mathrm{d}\mu_i(x)\,.
\end{align}
Unfortunately, such maps $T_i$ are not given, since we are only provided unpaired samples before $\mu_i$ and after $\nu_i$ that map's application.
By \citeauthor{Brenier1987}'s theorem, we know, however, that such an OT map $T^\star_i$ exists, and that it would be necessarily the gradient of a convex potential function that maximizes \eqref{eq:dual}. As a result, we propose to modify \eqref{eq:newobj1} to (i.) parameterize, for any $c$, the map $\mathcal{T}_\theta(c)$ as the gradient w.r.t. $x$ of a function $f_\theta(x,c):\mathbb{R}^d\times \mathcal{C}\rightarrow \mathbb{R}$ that is convex w.r.t. $x$, namely $\mathcal{T}_\theta(\cont) := x \mapsto \nabla_1 f_\theta(x,\cont)$; (ii.) estimate $\theta$ by maximizing \textit{jointly} the dual objectives~\eqref{eq:dual} simultaneously for all $N$ pairs of measures, in order to ensure that the maps are close to optimal, to form the aggregate problem
\begin{align}\label{eq:supdual} 
\textstyle \max_\theta \sum_{i=1}^N \mathcal{E}_{\mu_i,\nu_i}(f_{\theta}(\,\cdot\,, c_i)).
\end{align}
We detail in App.~\ref{app:algorithm} how the Legendre transforms that appear in the energy terms $\mathcal{E}_{\mu_i,\nu_i}$ are handled with an auxiliary function.

\subsection{Integrating Context in Convex Architectures}
We propose to incorporate context variables, in order to modulate a family of convex functions $f_{\theta}(x, c)$
using partially input convex neural networks (PICNN). PICNNs are neural networks that can be evaluated over a pair of inputs $(x,\cont)$, but which are only required to be convex w.r.t.~$x$. Given an input vector $x$ and context vector $\cont$, a $K$-layer PICNN is defined as $\psi_\theta(x, c) = z_{K},$ where, recursively for $0\leq k \leq K-1$ one has
\begin{equation} \label{eq:picnn}
\begin{aligned} 
u_{k+1} =&\, \tau_{k}\left(V_{k} u_{k}+v_{k}\right), \\
z_{k+1} =&\, \sigma_{k}\left(W_{k}^{z}\left(z_{k} \circ\left[W_{k}^{z u} u_{k}+b_{k}^{z}\right]_+\right)+\right.
\left.W_{k}^{x}\left(x \circ(W_{k}^{x u} u_{k}+b_{k}^{x})\right)+W_{k}^{u} u_{k}+b_{k}^u\right), \\
\end{aligned}
\end{equation}
where the PICNN is initialized as $u_{0}=\cont, z_0 = \mathbf{0}$, $\circ$ denotes the Hadamard elementwise product, and $\tau_k$ is any activation function. The parameters of the PICNN are then given by
$$\theta = \{ V_{k}, W_{k}^{z}, W_{k}^{z u}, W_{k}^{x}, W_{k}^{x u} , W_{k}^{u}, v_{k}, b_{k}^{z}, b_{k}^{x}, b_{k}^u \}.$$ 
Similar to ICNNs, the convexity w.r.t. input variable $x$ is guaranteed as long as activation functions $\sigma_i$ are convex and non-decreasing, and the weight matrices $W_{k}^{z}$ have non-negative entries. We parameterize this by storing them as elementwise applications of softplus operations on precursor matrices of the same size, or, alternatively, by regularizing their negative part. Finally, much like ICNNs, all matrices at the $K-1$ layer are line vectors, and their biases scalars.

Such networks were proposed by \citet[Eq. 3]{amos2017input} to address a problem that is somewhat symmetric to ours: Their inputs were labeled as $(y, x)$, where $y$ is a label vector, typically much smaller than that of vector $x$. Their PICNN is convex w.r.t.~$y$, in order to easily recover, given a datapoint $x$ (e.g., an image) the best label $y$ that corresponds to $x$ using gradient descent as a subroutine, i.e. $y^\star(x) = \arg\min_y \text{PICNN}_\theta(x,y)$. PICNN were therefore originally proposed to learn a parameterized, implicit classification layer, amortized over samples, whose motivation rests on the property that it is convex w.r.t. label variable $y$. By contrast, we use PICNNs that are convex w.r.t. data points $x$. In addition to that swap, we do not use the convexity of the PICNN to define an implicit layer (or to carry out gradient descent). Indeed, it does not make sense in our setting to minimize $\psi_\theta(x,\cont)$ as a function of $x$, since $x$ is an observation. Instead, our work rests on the property that $\nabla_1 \psi_\theta(x,\cont)$ describes a parameterized family of OT maps. We note that PICNNs were considered within the context of OT in \cite[Appendix B]{fan2021scalable}. In that work, PICNN provide an elegant reformulation for neural Wasserstein barycenters. \citet{fan2021scalable} considered a context vector $c$ that was restricted to be a small vector of probabilities.

\subsection{Conditional Monge Map Architecture}\label{subsec:combin}
Using PICNNs as a base module, the \textsc{CondOT} architecture integrates operations on the contexts $\Ccal$. As seen in Figure~\ref{fig:overview}, context values $\cont$ may take various forms:
\begin{enumerate}[noitemsep,leftmargin=.35cm,topsep=0pt,parsep=0pt,partopsep=0pt]
\item A scalar $t$ denoting a strength or a temporal effect. \newtext{For instance,} \citeauthor{mccann1997convexity}'s interpolation and its time parameterization, $\alpha_{t}=((1-t) \text{Id}+t T)_{\sharp} \alpha_{0}$ \citep{mccann1997convexity} can be interpreted as a \newtext{trivial conditional OT model that creates, from an OT map $\psi_\theta$, a set of maps parameterized by $t$}, $\mathcal{T}_\theta(t):=x\mapsto\nabla_x \left((1-t)\|x\|^2/2 +t \psi_\theta(x)\right)$.
\item A covariate vector influencing the nature of the effect that led $\mu_i$ to $\nu_i$, (capturing, e.g., patient feature vectors).
\item One or multiple actions, possibly discrete, representing decisions or perturbations applied onto $\mu_i$.
\end{enumerate}

To provide a flexible architecture capable of modeling different types of conditions as well as conditions appearing in combinations, the more general \textsc{CondOT} architecture consists of the hypernetwork $\mathcal{T}_\theta$ that is fed a context vector through embedding and combinator modules. This generic architecture provides a one-size fits all approach to integrate all types of contexts $\cont$.

\paragraph{Embedding Module}
To give greater flexibility when setting the context variable $c$, \textsc{CondOT} contains an embedding module $\mathcal{E}$ that translates arbitrary contexts into real-valued vectors. Besides simple scalars $t$ (Fig.~\ref{fig:overview}a) for which no embedding is required, discrete contexts can be handled with an embedding module $\mathcal{E}_\phi$.
When the set $\Ccal$ is small, this can be done effectively using one-hot embeddings $\mathcal{E}_\text{ohe}$.
For more complicated actions $a$ such as treatments, there is no simple way to vectorize a context $c$. Similarly to action embeddings in reinforcement learning \citep{chandak2019learning, tennenholtz2019natural}, we can learn embeddings for discrete actions into a learned continuous representation.
This often requires domain-knowledge on the context values. For molecular drugs, for example, we can learn molecular representations $\mathcal{E}_\text{mol}$ such as chemical, motif-based~\citep{rogers2010extended} or neural fingerprints~\citep{rong2020grover, schwaller2022machine}.
However, often this domain knowledge is not available.
In this work, we thus construct so-called \emph{mode-of-action} embeddings, by computing an embedding $\mathcal{E}_\text{moa}$ that encourages actions $a$ with similar effect on target population $\nu$ to have a similar representation.
In \S~\ref{sec:evaluation}, we analyze several embedding types for different use-cases.

\paragraph{Combinator Module}
\begin{wrapfigure}{r}{0.4\textwidth}
    \centering
    \includegraphics[width=.4\textwidth]{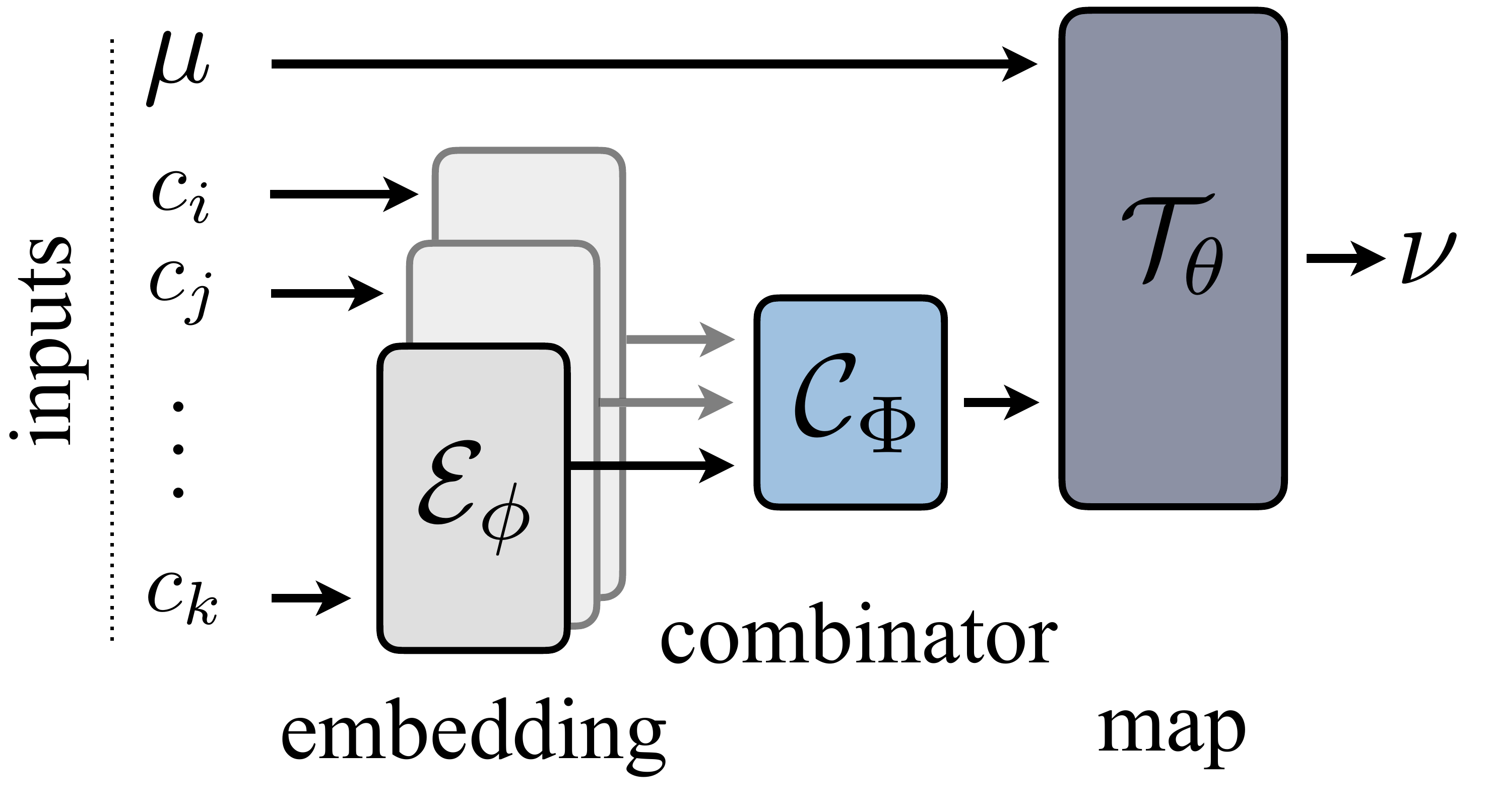}
    \caption{\textbf{\textsc{CondOT} Architecture and Modules}. The embedding module $\mathcal{E}_\phi$ embeds arbitrary conditions $c$, which are then combined via module $\mathcal{C}_\Phi$. Using the processed contexts $c$, the map $\mathcal{T}_\theta(c)$ acts on $\mu$ to predict the target measure $\nu$.}\vskip-0.5cm
    \label{fig:architecture}
\end{wrapfigure}
\looseness -1 While we often have access to contexts $c$ in isolation, it is crucial to infer the effect of contexts applied in combination. A prominent example are cancer combination therapies, in which multiple treatment modalities are administered in combination to enhance treatment efficacy~\citep{kummar2010utilizing}.
In these settings, the mode of operation between individual contexts $c$ is often not known, and can thus not be directly modeled via simple arithmetic operations such as 
\texttt{min, max, sum, mean}.
While we test as a baseline the case, applicable to one-hot-embeddings, where simple additions are used to model these combinations, we propose to augment the \textsc{CondOT} architecture with a parameterized combinator module $\mathcal{C}_\Phi$.
If the order in which the actions are applied is irrelevant or unknown, the corresponding network $\mathcal{C}_\Phi$ needs to be permutation-invariant,
which can be achieved by using a deep set architecture~\citep{zaheer2017}.
Receiving a flexible number of inputs from the embedding module $\mathcal{E}_\phi$, \textsc{CondOT} allows for a joint training of the PICNN parameters $\theta$, embedding parameters $\phi$, and combinator parameters $\Phi$ in a single, end-to-end differentiable architecture.

\paragraph{Training Procedure}
\looseness -1 Given a dataset $\mathcal{D}=\{c_i, (\mu_i, \nu_i) \}_{i=0}^N$ of $N$ pairs of populations before $\mu_i$ and after transport $\nu_i$ connected to a context $c_i$, we detail in Algorithm~\ref{algo:condot} provided in \S~\ref{app:algorithm}, a training loop that incorporates all of the architecture proposals described above. The training loss aims at making sure the map $\mathcal{T}_\theta(c_i)$ is an OT map from $\mu_i$ to $\nu_i$, where $c_i$ may either be the original label itself or its embedded/combined formulation in more advanced tasks. To handle the Legendre transform in \eqref{eq:dual}, we use the proxy dual objective defined in \citep[Eq. 6]{makkuva2020optimal} \eqref{eq:makkuva_f_loss}-\eqref{eq:makkuva_g_loss} in place of \eqref{eq:dual} in our overall loss~\eqref{eq:supdual}.
This involves training the \textsc{CondOT} architecture using two PICNNs, i.e., $\text{PICNN}_{\theta_f}$ and $\text{PICNN}_{\theta_g}$, that share the same embedding/combinator module, with a regularization \eqref{eq:ot-minmax} promoting that for any $c$, the $\text{PICNN}_{\theta_g}(\cdot,c)$ resembles the Legendre transform of the other, $\text{PICNN}_{\theta_f}^*(\cdot,c)$.

\section{Initialization Strategies for Neural Convex Architectures} \label{sec:inits}
\looseness -1 We address the problem of initializing the parameters of (P)ICNNs to ensure their gradient evaluated at every point is (initially) meaningful in the context of OT, namely that it is able \newtext{to map the first and second moments} of a measure $\mu$ into \newtext{those of} a target measure $\nu$. 
\looseness -1 The initializers we propose build heavily on the \newtext{quadratic layers proposed in the seminal reference \citep[Appendix~B.2]{korotin2020wasserstein}, notably the ``DenseQuad'' layer, as well as} on closed-form solutions available for Gaussian approximations of measures~\citep{gelbrich1990formula}.

\paragraph{Closed-Form Potentials for Gaussians}
Given two Gaussian distributions $\mathcal{N}_1, \mathcal{N}_2$ with means respectively $\bm_1, \bm_2$ and covariance matrices $\Sigma_1, \Sigma_2$ (where $\Sigma_1$ is assumed to be full rank), the Brenier potential solving the OT problem from the first to the second Gaussian reads:
\begin{align} \label{eq:gaussian}
f^\star_{\mathcal{N}_1, \mathcal{N}_2} = \tfrac{1}{2}x^TA^TAx + b^T x + t(A,b) = \tfrac{1}{2}\|Ax\|_2^2 + b^T x + t(A,b),  \text{\,where,}\\
A := \left(\Sigma_1^{-1/2} \left(\Sigma_1^{1/2} \Sigma_2 \Sigma_1^{1/2}\right)^{1/2} \Sigma_1^{-1/2}\right)^{1/2},\;\; b := \bm_2 - A^T A \bm_1, \nonumber
\end{align}
define both quadratic and linear terms and $t(A,b)$ can be any constant. Importantly, note that we write the quadratic term in factorized form $AA^T$ to enforce psd-ness, \newtext{as done by~\cite{korotin2020wasserstein}}, not as usually done with a single psd matrix~\citep[Remark 2.31]{Peyre2019computational}.

Our quadratic potentials are only injected in the first state of hidden vector $z_0$, to populate it with a collection of relevant full-rank quadratic convex functions, with the goal of recovering \newtext{an affine OT map} from the start, as illustrated in the experiments from \S~\ref{app:comp_init}.

\paragraph{Quadratic Potentials Lower Bounded by 0} Naturally, for any choice of $t(A,b)$ one recovers the property that $\nabla f^\star_{\mu,\nu}\sharp\mathcal{N}_1= \mathcal{N}_2$. When used in deep architectures, the level of that constant does, however, play a role, since convex functions in ICNN are typically thresholded or modulated using rectifying functions. To remove this ambiguity, we settle on a choice for $t(A,b)$ that is such that the lowest value reached by $f^\star_{\mathcal{N}_1, \mathcal{N}_2}$ is 0. This can be obtained by setting
\begin{equation} \label{eq:gauss2}
t(A,b) := b^T (A^TA)^{-1} b\,,
\end{equation}
which results in the following choice, writing $\omega =\bm_1 - (A^TA)^{-1} \bm_2$,
\begin{equation}\label{eq:init}
f^\star_{\mathcal{N}_1, \mathcal{N}_2}(x) = \tfrac{1}{2}\|A\left(x + (A^TA)^{-1} b\right)\|_2^2 = \tfrac{1}{2}\|A\left(x - \omega\right)\|_2^2\,.
\end{equation}

\begin{figure*}[t]
    \centering
    \includegraphics[width=\textwidth]{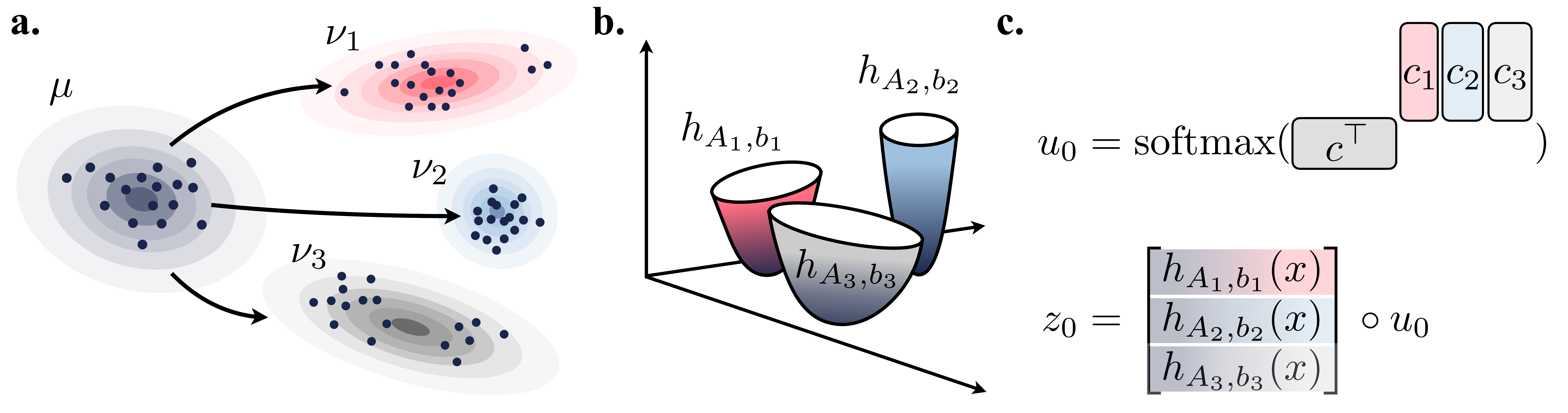}
    \caption{\textbf{a.} From a measure $\mu$ to several target measures $\nu_1, \nu_2, \nu_3$ provided with labels $c_1, c_2, c_3$ we can extract three Gaussian (quadratic) potentials in closed form, \textbf{b.} whose gradients transport on a first approximation $\mu$ to areas in space that cover the three targets. \textbf{c.} Given a new label vector $c$, we compare it to known labels to modulate the magnitude of each of the three potentials.}
    \label{fig:picnn}    
\end{figure*}

To mimic these potential functions, we introduce a quadratic \textit{layer} parameterized by a weight matrix $M$ and a ``bias'' vector $m$, defined as $q_{M,m}(x) = \tfrac{1}{2}\|M(x - m)\|_2^2$. By design, $q_{M,m}(x)$ is a convex quadratic, non-negative layer. Finally, one has the following relationships, 
\begin{equation}\label{eq:identities}
\nabla q_{I,\mathbf{0}_d}= \text{Id}, \quad \nabla q_{A,\omega}\sharp \mathcal{N}_1 = \mathcal{N}_2\;.
\end{equation}
\textbf{ICNN Initialization } 
We explore two possible ICNN \eqref{eq:icnn} initializers for OT.

\underline{Identity Initialization} The first approach ensures that upon initialization the ICNN's gradient mimics the \textit{identity} map, i.e., $\nabla \psi_{\theta}(x)=x$ for any $x$. We do so by injecting in the initial hidden state $z_0$ the norm of the input vector $\tfrac{1}{2}\|x\|^2$, cast as a trainable layer $q_{M,m}$ initialized with $M=I$ and $m=\mathbf{0}_d$, see~\eqref{eq:identities}. The remaining parameters are chosen to propagate that norm throughout layers using averages. This amounts to the following choices:
\begin{enumerate}[noitemsep,leftmargin=.35cm,topsep=0pt,parsep=0pt,partopsep=0pt]
\item Set all $\sigma_i$ to be activations such that $\sigma_i'(u)\approx 1$ for $u$ large enough, e.g., (leaky) ReLU or softplus.
\item Introduce an initialization layer, $z_0=q_{M,m}(x)\mathbf{1}$, itself initialized with $M=I$ and $m=\mathbf{0}_d$.
\item Initialize all matrices $W_i^z$ to $\approx \mathbf{1}_{d_2,d_1}/d_1$, where $d_1,d_2$ are the dimensions of these matrices.
\item Initialize all matrices $W^x_i$ to $\approx 0$.
\item Initialize biases $b_i$ to $s\mathbf{1}$, where $s$ is a large enough value $s$ so that $\sigma_i'(s)\approx 1$.
\end{enumerate}

\underline{Gaussian Initialization} The second approach can be used to initialize an ICNN so that its gradient mimics the affine transport between the Gaussian approximations of $\mu$ and $\nu$. To this end, we follow all of the steps outlined above, except for step 2 where the quadratic layer $q_{M,m}$ is initialized instead with $M=A$ and $m=\bm_1-(A^TA)^{-1}\bm_2$ using notations in~\eqref{eq:gaussian},~\eqref{eq:gauss2},~\eqref{eq:init}, where $\bm_1, \bm_2, \Sigma_1, \Sigma_2$ are replaced by the empirical mean and covariances of $\mu$ and $\nu$. \newtext{Throughout the experiments, we use the Gaussian and identity initialization. Further comparisons between the vanilla initialization and those introduced in this work can be found in \S~\ref{app:comp_init} (Fig.~\ref{fig:exp_comp_inits}).}

\paragraph{PICNN Initialization}
Recall for convenience that a $K$-layer PICNN architecture reads: $$
\begin{aligned}
u_{k+1} =&\, \tau_{k}\left(V_{k} u_{k}+v_{k}\right) \\
z_{k+1} =&\, \sigma_{k}\left(W_{k}^{z}\left(z_{k} \circ\left[W_{k}^{z u} u_{k}+b_{k}^{z}\right]_+\right)+\right.
\left.W_{k}^{x}\left(x \circ\left(W_{k}^{x u} u_{k}+b_{k}^{x}\right)
\right)+W_{k}^{u} u_{k}+b_{k}^u\right) \\
\psi_\theta(x, c) =&\, z_{K}.
\end{aligned}
$$
In their original form~\citep[Eq. 3]{amos2017input}, PICNNs are initialized by setting $u_0=c$ and $z_0=\mathbf{0}$ to a zero vector of suitable size. Intuitively, the hidden states $u_k$ act as context-dependent modulators, whereas vectors $z_k$ propagate, layer after layer, a collection of convex functions in $x$ that are iteratively refined, while retaining the property that they are each convex in $x$. A reasonable initialization for a PICNN that is provided a context vector $c$ is that if $c\approx c_j$ (where $j$ is in the training set), one has that $\nabla_1\psi_{\theta_0}(\cdot,c)$ maps approximately $\mu_j$ to $\nu_j$, which can be obtained by having $\psi_{\theta_0}(\cdot, c)$ mimic the closed-form Brenier potential between the Gaussian approximations of $\mu_j, \nu_j$. Alternatively, one may also default to an identity initialization as discusses above. To obtain either behavior, we make the following modifications, and refer to the illustration in Fig.~\ref{fig:picnn}:

\begin{enumerate}[noitemsep,leftmargin=.35cm,topsep=0pt,parsep=0pt,partopsep=0pt]
\item The modulator $u_0(c)=\text{softmax}(c^TM)$, where $C$ is initialized as $M=[c_j]_j$, and $V_i=I, v_i=\mathbf{0}$.
\item $z_0=[q_{M_j, m_j}(x)]_j$, where weight matrices and bias $(M_j, m_j)$ are either initialized to $(I,\mathbf{0})$ or as $(A_j, \omega_j)$ recovered by solving the Gaussian affine map from $\mu_j$ to $\nu_j$ using ~\eqref{eq:init}.
\item Modulator $u_0$ is passed directly to hidden state upon first iteration $W_0^{zu}=I, b_0^z=0$.
\item All subsequent matrices $W_k^z$ are initialized to $\approx \mathbf{1}_{d_2,d_1}/d_1$, where $d_1,d_2$ are their dimensions,
\item $W^x_k$ and $W^{xu}_k$ are $\approx 0$, the biases $b^z_k\approx\mathbf{1}$, $b_k^u\approx\mathbf{0}$.
\end{enumerate}

\begin{table*}[t]
    \caption{Evaluation of drug effect predictions from control cells to cells treated with drug Givinostat when conditioning on various covariates influencing cellular responses such as drug dosage and cell type. Results are reported based on MMD and the $\ell_2$ distance between perturbation signatures of marker genes in the 1000 dimensional gene expression space.}
    \label{tab:exp_scalar_covariate_sciplex}
    \centering
\adjustbox{max width=\linewidth}{%
    \begin{tabular}{lcccccccc}
    \toprule
         \textbf{Method} & \multicolumn{6}{c}{\textbf{Conditioned on Drug Dosage}} & \multicolumn{2}{c}{\textbf{Conditioned on Cell Line}} \\
         & \multicolumn{2}{c}{In-Sample} && \multicolumn{2}{c}{Out-of-Sample} && \multicolumn{2}{c}{In-Sample} \\
         \cmidrule{2-3} \cmidrule{5-6} \cmidrule{8-9}
         & MMD & $\ell_2(\text{PS})$ && MMD & $\ell_2(\text{PS})$ && MMD & $\ell_2(\text{PS})$ \\
    \midrule
        \textsc{CPA} \citep{lotfollahi2021compositional} & 0.1502 $\pm$ 0.0769 & 2.47 $\pm$ 2.89 && 0.1568 $\pm$ 0.0729 & 2.65 $\pm$ 2.75 && 0.2551 $\pm$ 0.006 & 2.71 $\pm$ 1.51 \\
        \textsc{ICNN OT} \citep{makkuva2020optimal} & 0.0365 $\pm$ 0.0473 & 2.37 $\pm$ 2.15 && 0.0466 $\pm$ 0.0479 & 2.24 $\pm$ 2.39 && 0.0206 $\pm$ 0.0109 & 1.16 $\pm$ 0.75 \\
        \textsc{CondOT} (Identity initialization) & 0.0111 $\pm$ 0.0055 & 0.63 $\pm$ 0.09 && 0.0374 $\pm$ 0.0052 & 2.02 $\pm$ 0.10 && 0.0148 $\pm$ 0.0078 & 0.39 $\pm$ 0.06 \\
        \textsc{CondOT} (Gaussian initialization) & 0.0128 $\pm$ 0.0081 & 0.60 $\pm$ 0.11 && 0.0325 $\pm$ 0.0062 & 1.84 $\pm$ 0.14 && 0.0146 $\pm$ 0.0074 & 0.41 $\pm$ 0.07 \\
    \bottomrule
    \vspace{-20pt}
    \end{tabular}
}
\end{table*}

\begin{figure*}
    \centering
    \includegraphics[width=1.2\textwidth]{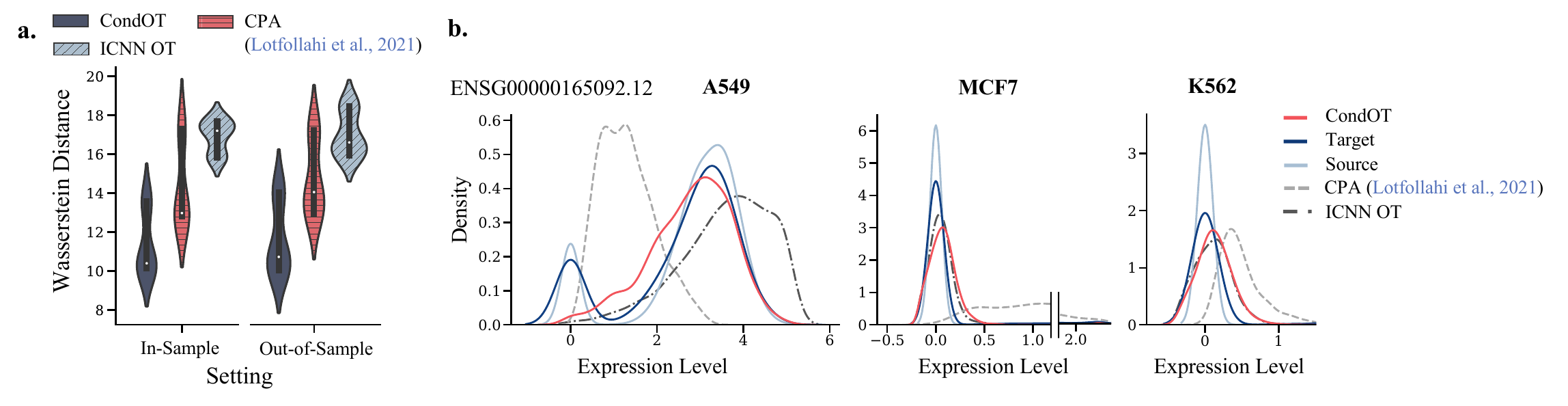}
    \caption{\textbf{a.} \looseness -1 Predictive performance of \textsc{CondOT} and baselines w.r.t. the entropy-regularized Wasserstein distance on drug dosages \emph{in-sample}, i.e., seen during training, and \emph{out-of-sample}, i.e., unseen during training. \textbf{b.} Marginal distributions of observed source and target distributions, as well as predictions on perturbed distributions by \textsc{CondOT} and baselines of an exemplary gene across different cell lines. Predicted marginals of each method should match the marginal of the target population.}
    \label{fig:exp_scalar_sciplex}
\end{figure*}

\vspace{-5pt}
\section{Evaluation} \label{sec:evaluation}
\vspace{-5pt}

\looseness -1 Biological cells undergo changes in their molecular profiles upon chemical, genetic, or mechanical perturbations. These changes can be measured using recent technological advancements in high-resolution multivariate single-cell biology. Measuring single cells in their unperturbed or perturbed state requires, however, to destroy them, resulting in populations $\mu$ and $\nu$ that are unpaired. The relevance of OT to that comes from its ability to resolve such ambiguities through OT maps, 
holding promises of a better understanding of health and disease. 
We consider various high-dimensional problems arising from this scenario to evaluate the performance of \textsc{CondOT} (\S~\ref{sec:method}) versus other baselines.

\subsection{Population Dynamics Conditioned on \emph{Scalars}} \label{sec:eval_scalar}

\looseness -1 Upon application of a molecular drug, the state of each cell $x_i$ of the unperturbed population is altered, and observed in population $\nu$.
Molecular drugs are often applied at different dosage levels $t$, and the magnitude of changes in the gene expression profiles of single cells highly correlates with that dosage. 
We seek to learn a global, parameterized transport map $\mathcal{T}_\theta$ sensitive to that dosage.
We evaluate our method on the task of inferring single-cell perturbation responses to the cancer drug Givinostat, a histone deacetylase inhibitor with potential anti-inflammatory, anti-angiogenic, and antineoplastic activities \citep{srivatsan2020massively}, applied at different dosage levels, i.e., $t \in \{10\,$nM, $100\,$nM, $1,000\,$nM, $10,000\,$nM$\}$. The dataset contains $3,541$ cells described with the gene expression levels of $1,000$ highly-variable genes.
In a first experiment, we measure how well \textsc{CondOT} captures the drug effects at different dosage levels via distributional distances such as MMD~\citep{gretton2012kernel} and the $\ell_2$-norm between the corresponding perturbation signatures (PS), as well as the entropy-regularized Wasserstein distance~\citep{cuturi2013sinkhorn}. We compute the metrics on 50 marker genes, i.e., genes mostly affected upon perturbation. For more details on evaluation metrics, see \S~\ref{app:eval_metrics}. To put \textsc{CondOT}'s performance into perspective, we compare it to current state-of-the-art baselines~\citep{lotfollahi2021compositional} as well as parameterized Monge maps without context variables \citep[\textsc{ICNN OT}]{bunne2021learning, makkuva2020optimal}, see \S~\ref{app:baselines}.
As visible in Table~\ref{tab:exp_scalar_covariate_sciplex} and Fig.~\ref{fig:exp_scalar_sciplex}a, \textsc{CondOT} achieves consistently more accurate predictions of the target cell populations at different dosage levels than OT approaches that cannot utilize context information, demonstrated through a lower average loss and a smaller variance.
This becomes even more evident when moving to the setting where the population has been trained only on a subset of dosages and we test \textsc{CondOT} on \emph{out-of-sample} dosages. Table~\ref{tab:exp_scalar_covariate_sciplex} and Fig.~\ref{fig:exp_scalar_sciplex}a demonstrate that \textsc{CondOT} is able to generalize to previously \emph{unknown} dosages, thus learning to interpolate the perturbation effects from dosages seen during training. For further analysis, we refer the reader to \S~\ref{app:exp_details} (see Fig.~\ref{fig:exp_scalar_sciplex_marginals} and \ref{fig:exp_scalar_sciplex_umaps}). We further provide an additional comparison of \textsc{CondOT}, operating in the multi-task setting, to the single-task performance of optimal transport-based methods \S~\ref{app:add_baseline}. While the single-task setting of course fails to generalize to new contexts and requires all contexts to be distinctly known, it provides us with a \textit{pseudo} lower bound, which \textsc{CondOT} is able to reach (see Table~\ref{tab:lower_bound}).

\begin{figure*}[t]
    \includegraphics[width=\textwidth]{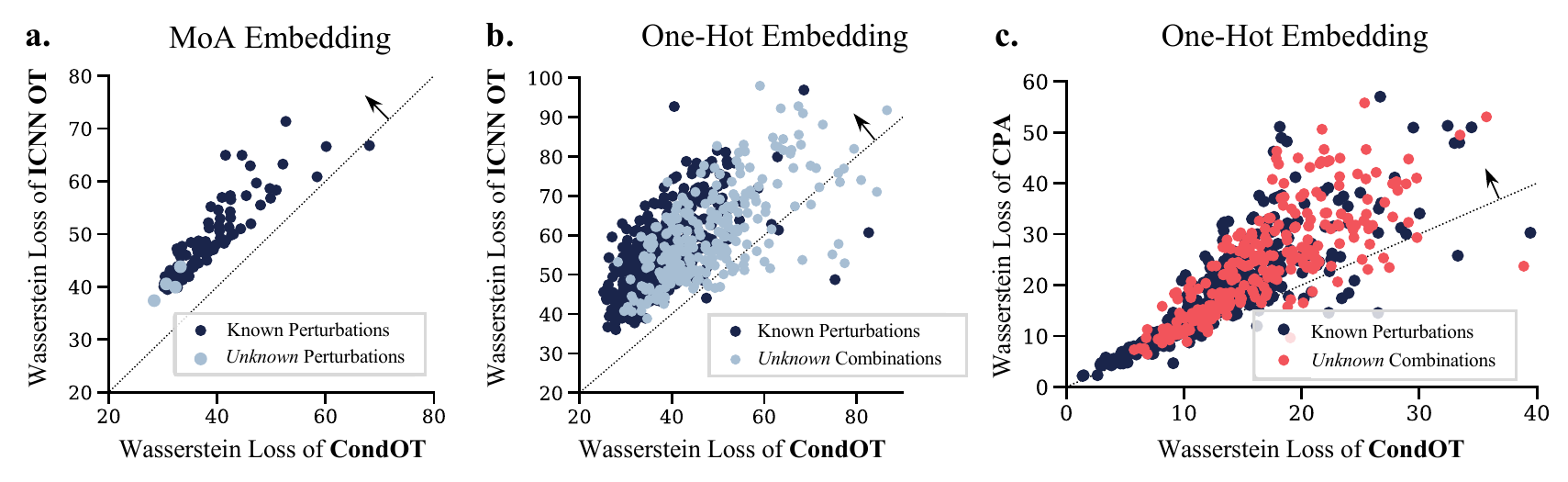}
    \caption{Comparison between \textbf{a.} \textsc{CondOT} and \textsc{ICNN OT}~\citep{makkuva2020optimal} based on embedding $\mathcal{E}_\text{moa}$ \textbf{b.} as well as $\mathcal{E}_\text{ohe}$, and \textbf{c.} \textsc{CondOT} and \textsc{CPA}~\citep{lotfollahi2021compositional} based on embedding $\mathcal{E}_\text{ohe}$ on \emph{known} and \emph{unknown} perturbations or combinations. Results above the diagonal suggest higher predictive performance of \textsc{CondOT}.}
    \label{fig:exp_action_norman_scatter}
\end{figure*}

\subsection{Population Dynamics Conditioned on \emph{Covariates}} \label{sec:eval_covariate}

\looseness -1 Molecular processes are often highly dependent on additional covariates that steer experimental conditions, and which are not present in the features measures in population $\mu$ or $\nu$.
This can be, for instance, factors such as different cell types clustered within the populations.
When the model can only be conditioned w.r.t. a small and \textit{fixed} set of metadata information, such as cell types, it is sufficient to encode these contexts using a one-hot embedding module $\mathcal{E}_\text{ohe}$.
To illustrate this problem, we consider cell populations comprising three different cell lines (A549, MCF7, and K562). As visible in Table~\ref{tab:exp_scalar_covariate_sciplex}, \textsc{CondOT} outperforms current baselines which equally condition on covariate information such as \textsc{CPA}~\citep{lotfollahi2021compositional}, assessed through various evaluation metrics.
Figure~\ref{fig:exp_scalar_sciplex}b displays a gene showing highly various responses towards the drug Givinostat dependent on the cell line. \textsc{CondOT} captures the distribution shift from control to target populations consistently across different cell lines.

\subsection{Population Dynamics Conditioned on \emph{Actions}}

\looseness -1 To recommend personalized medical procedures for patients, or to improve our understanding of genetic circuits, it is key to be able to predict the outcomes of novel perturbations, arising from combinations of drugs or of genetic perturbations. 
Rather than learning individual maps $T_\theta^a$ predicting the effect of individual treatments, we aim at learning a global map  $\mathcal{T}_\theta$ which, given as input the unperturbed population $\mu$ as well as the action $a$ of interest, predicts the cell state perturbed by $a$.
Thanks to its modularity, \textsc{CondOT} can not only learn a map $T_\theta$ for all actions \emph{known} during training, but also to generalize to \emph{unknown} actions, as well as potential \emph{combinations} of actions. We will discuss all three scenarios below.

\subsubsection{\textit{Known} Actions} \label{sec:eval_action_known}
\vspace{-5pt}

\begin{figure*}[t]
    \includegraphics[width=1.15\textwidth]{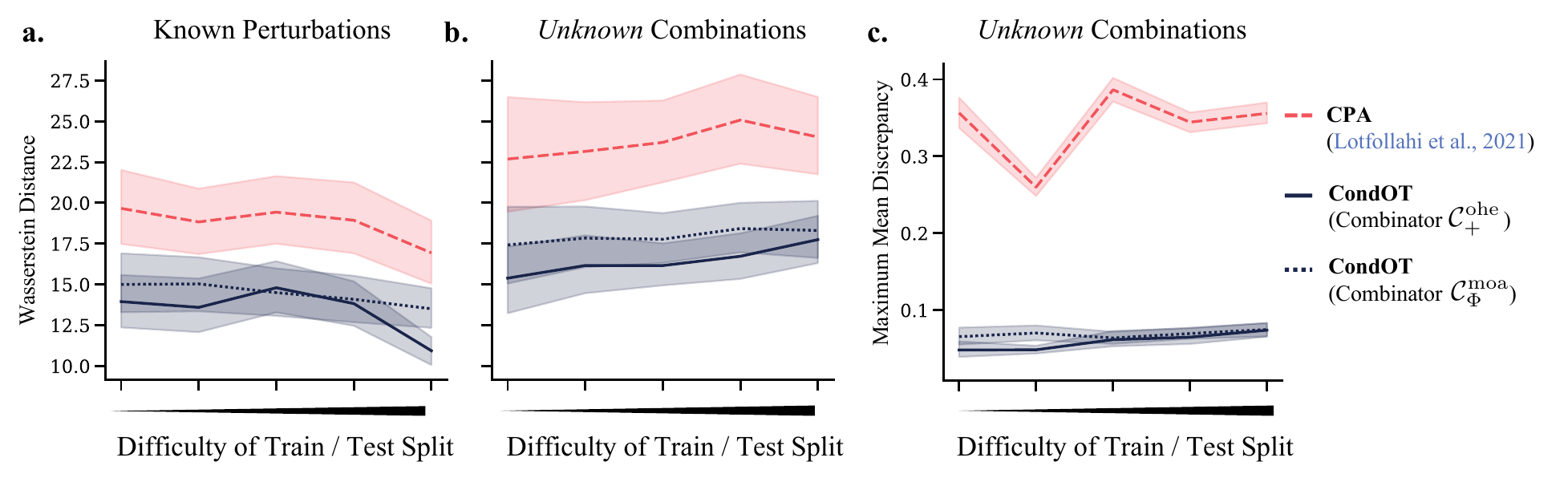}
    \caption{Predictive performance for \textbf{a.} known perturbations, \textbf{b.} unknown perturbations in combination w.r.t. regularized Wasserstein distance and \textbf{c.} MMD over different train / test splits of increasing difficulty for baseline \textsc{CPA} as well as \textsc{CondOT} with different combinators $\mathcal{C}^\text{ohe}_+$ and $\mathcal{C}^\text{moa}_\Phi$. For more details on the dataset splits, see~\S\ref{app:datasplits}.}
    \label{fig:exp_action_norman_line}
\end{figure*}

\looseness -1 In the following, we analyze \textsc{CondOT}'s ability to accurately predict phenotypes of genetic perturbations based on single-cell RNA-sequencing pooled CRISPR screens \citep{norman2019exploring, dixit2016perturb}, comprising $98,419$ single-cell gene expression profiles with $92$ different genetic perturbations, each cell measured via a $1,500$ highly-variable gene expression vector.
As, in a first step, we do not aim at generalizing beyond perturbations encountered during training, we utilize again a one-hot embedding $\mathcal{E}_\text{ohe}$ to condition $\mathcal{T}_\theta$ on each perturbation $a$.
We compare our method to other baselines capable of modeling effects of a large set of perturbations such as \textsc{CPA} \citep{lotfollahi2021compositional}.
Often, the effect of genetic perturbations are subtle in the high-dimensional gene expression profile of single cells. Using ICNN-parameterized OT maps without context information, we can thus assess the gain in accuracy of predicting the perturbed target population by incorporating context-awareness over simply predicting an average perturbation effect. 
Figure~\ref{fig:exp_action_norman_scatter}a and b demonstrate that compared to OT ablation studies, Fig.~\ref{fig:exp_action_norman_scatter}c and Fig.~\ref{fig:exp_action_norman_line}a for the current state-of-the-art method \textsc{CPA}~\citep{lotfollahi2021compositional}. Compared to both, \textsc{CondOT} captures the perturbation responses more accurately w.r.t. the Wasserstein distance.

\subsubsection{\textit{Unknown} Actions} \label{sec:eval_action_unknown}
\vspace{-5pt}

\looseness -1 With the emergence of new perturbations or drugs, we aim at inferring cellular responses to settings not explored during training.
One-hot embeddings, however, do not allow us to model \emph{unknown} perturbations. 
This requires us to use an embedding $\mathcal{E}$, which can provide us with a representation of an unknown action $a^\prime$.
As genetic perturbations further have no meaningful embeddings as, for example, molecular fingerprints for drugs, we resort to mode-of-action embeddings introduced in \S~\ref{subsec:combin}. Assuming marginal sample access to all individual perturbations, we compute a multidimensional scaling (MDS)-based embedding from pairwise Wasserstein distances between individual target populations, such that perturbations with similar effects are closely represented. For details, see \S~\ref{app:exp_details}.
As current state-of-the-art methods are restricted to modeling perturbations via one-hot encodings, we compare our method to \textsc{ICNN OT} only. As displayed in Fig.~\ref{fig:exp_action_norman_scatter}a, \textsc{CondOT} accurately captures the response of \emph{unknown} actions (BAK1, FOXF1, MAP2K6, MAP4K3), which were not seen during training, at a similar Wasserstein loss as perturbation effects seen during training. For more details, see \S~\ref{app:exp_details}.
 
 \begin{figure*}
    \centering
    \includegraphics[width=0.8\textwidth]{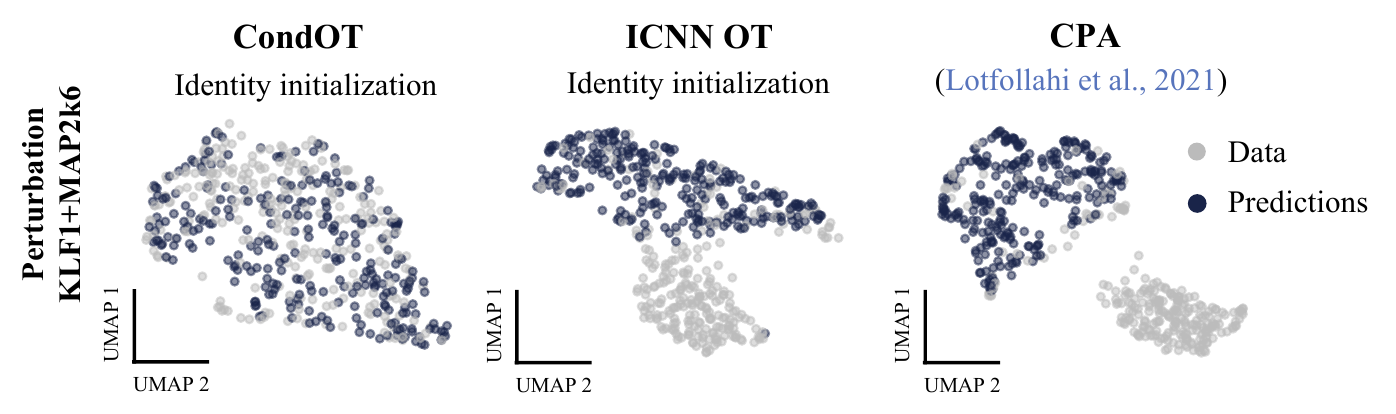}
    \caption{\looseness -1 UMAP embeddings of cells perturbed by the combination KLF1+MAP2K6 (gray) and predictions of \textsc{CondOT} (ours), \textsc{ICNN OT}~\citep{makkuva2020optimal}, and \textsc{CPA} (blue). While \textsc{CondOT} aligns well with observed perturbed cells, the baselines fail to capture subpopulations.}
    \label{fig:action_comb_comparison_umap}
\end{figure*}

\subsubsection{Actions in Combination} \label{sec:eval_action_comb}
\vspace{-5pt}

\looseness -1 While experimental studies can often measure perturbation effects in biological systems in isolation, the combinatorial space of perturbations in composition is too large to capture experimentally. Often, however, combination therapies are cornerstones of cancer therapy~\citep{mokhtari2017combination}.
In the following, we test different combinator architectures to predict genetic perturbations in combination from single targets.
Similarly to \citet{lotfollahi2021compositional}, we can embed combinations by adding individual one-hot encodings of single perturbations (i.e., $\mathcal{C}^\text{ohe}_+$). In addition, we parameterize a combinator via a permutation-invariant deep set, as introduced in \S~\ref{subsec:combin}, based on mode-of-action embeddings of individual perturbations (i.e., $\mathcal{C}^\text{moa}_\Phi$). 
We split the dataset into train / test splits of increasing difficulty (details on the dataset splits in~\S\ref{app:datasplits}). Initially containing all individual perturbations as well as some combinations, the number of perturbations seen in combination during training decreases over each split. For more details, see \S~\ref{app:exp_details}.
We compare different combinators to \textsc{ICNN OT} (Fig.~\ref{fig:exp_action_norman_scatter}b) and \textsc{CPA}~\citep{lotfollahi2021compositional} (Fig.~\ref{fig:exp_action_norman_scatter}c, Fig.~\ref{fig:exp_action_norman_line}b, c). While the performance drops compared to inference on \emph{known} perturbations (Fig.~\ref{fig:exp_action_norman_line}a) and decreases with increasing difficulty of the train / test split, \textsc{CondOT} outperforms all baselines.
When embedding these high-dimensional populations in a low-dimensional UMAP space~\citep{umap}, one can see that \textsc{CondOT} captures the entire perturbed population, while \textsc{ICNN OT} and \textsc{CPA} fail in capturing certain subpopulations in the perturbed state (see Fig.~\ref{fig:action_comb_comparison_umap} and ~\ref{fig:exp_scalar_norman_umaps_more}).

\section{Conclusion}
We have developed the \textsc{CondOT} framework that is able to infer OT maps from not only one pair of measures, but many pairs that come labeled with a context value. To ensure that \textsc{CondOT} encodes optimal transports, we parameterize it as a PICNN, an input-convex NN that modulates the values of its weights matrices according to a sequence of feature representations of that context vector. We showcased the generalization abilities of \textsc{CondOT} in the extremely challenging task of predicting outcomes for unseen combinations of treatments. These abilities and PICNN more generally hold several promises, both as an augmentation of the \texttt{OTT} toolbox~\citep{cuturi2022optimal}, and for future applications of OT to single-cell genomics.

\begin{ack}
This publication was supported by the NCCR Catalysis (grant number 180544), a National Centre of Competence in Research funded by the Swiss National Science Foundation. We thank Stefan Stark and Gabriele Gut for helpful discussions and the reviewers for their thoughtful comments and efforts towards improving our manuscript.
\end{ack}

\bibliographystyle{abbrvnat}
\bibliography{main}

\newpage
\newpage

\section*{Appendix}
\appendix

\section{Background}

\paragraph{Primal and Dual Optimal Transport} The primal optimal transport problem (POT) was introduced in \eqref{eq:monge}, and quickly linked in our background section \S~\ref{sec:background} to the dual optimal transport problem (DOT) \eqref{eq:dual}. We provide for completeness an intermediary step to facilitate understanding, which works in the case where $p=2$, and explain why the optimal transport map $T$ can also be recovered via the dual optimal transport problem. Introduced by \citeauthor{kantorovich1942transfer} in \citeyear{kantorovich1942transfer}, the dual formulation is a constrained concave maximization problem defined as
\begin{equation*} \label{eq:kantorovich}
    W(\mu, \nu)= \sup _{(f, g) \in \Psi_c} \int_{\mathcal{X}} f(x) \mathrm{d} \mu(x)+\int_{\mathcal{Y}} g(y) \mathrm{d} \nu(y),
\end{equation*}
where the set of admissible potentials is $\Psi_c \defeq \{(f, g) \in L^{1}(\mu) \times L^{1}(\nu): f(x)+g(y) \leq \frac{1}{2}\|x-y\|$, $\forall(x, y) d\mu \otimes d\nu \text{ a.e.}\}$ \citep[Theorem 1.3]{villani2021topics}. The machinery of $c$-transforms~\citep[\S1.3 1]{santambrogio2015optimal} can be used to simplify that problem. When the cost $c$ is half the square Euclidean distance as considered here, this results in a simpler, so-called semi-dual problem~\citep{semidual} that only involves a single potential function.
\begin{equation}\label{eq:dual2}
f^\star_{\mu,\nu}:=\arg\sup_{f\,\text{convex}}\int_{\mathbb{R}^d}f^*\textrm{d}\mu+\int_{\mathbb{R}^d}f\textrm{d}\nu\,= \bpot^\star_{\mu,\nu}\,.
\end{equation} The optimal convex potential function $\bpot$ is then related to the optimal dual potential $f^\star$ expressed above, through the identity $\bpot = f^\star_{\mu,\nu}$.

\paragraph{Neural Optimal Transport} 
Learning optimal transport problems based on neural networks is at the core of many machine learning applications, including normalizing flows \citep{rezende2015variational,huang2021convex} and generative models \citep{arjovsky2017wasserstein,genevay2018}.
Directly parameterizing the doubly-stochastic matrix $T$ of the primal optimal transport \eqref{eq:monge} as done in previous work~\citep{jacob2018w2gan, yang2018scalable, prasad2020optimal} has been shown to yield an unstable and thus difficult to solve optimization problem~\citep[Table 1]{makkuva2020optimal}.
We thus follow previous work \citep{makkuva2020optimal,bunne2022proximal,korotin2020wasserstein, alvarez2021optimizing} and instead learn map $T$ via the convex Brenier potential $\bpot$ connected to the primal and dual optimal transport problem as outlined above.
We parameterize the convex function $\bpot$ via convex neural architectures (see \S~\ref{sec:icnn}), which can thus be used in two contexts, either to model the Brenier potential, or to model a dual function. Both lead to the same results since the Brenier potential $\psi^\star_{\mu,\nu}$ is equal to the optimal dual potential associated with the second measure $\nu$, $f^\star_{\mu,\nu}$, as described above and in \S\ref{sec:background} around~\eqref{eq:dual}. 

\section{The \textsc{CondOT} Algorithm} \label{app:algorithm}

\textsc{CondOT} provides a generalized approach that from \emph{labeled} pairs of measures $\{(\cont_i, (\mu_i, \nu_i))\}_i$ infers a \textit{global} parameterized conditional Monge map $\mathcal{T}_{\theta}$.
This is achieved by jointly learning an the embedding module $\mathcal{E}_\phi$, a combinator module $\mathcal{C}_\Phi$, as well as transport map $\mathcal{T}_{\theta}$. The algorithmic procedure is outlined in Algorithm~\ref{algo:condot}. We describe \textsc{CondOT}'s modules as well as their parameterization in detail in~\S\ref{app:modules}. In the following, we will cover in more depth algorithmic approaches on how to learn transport map $\mathcal{T}_{\theta}$.
Several approaches have been proposed on inferring transport map $\mathcal{T}_{\theta}$ from paired source and target populations, including the primal \eqref{eq:monge} or dual optimal transport problem \eqref{eq:dual}.

A possible approach to learn our model could consist in minimizing a primal OT problem. In that case, we can learn $\mathcal{T}_{\theta}$ via the gradient of the Brenier potential parameterized via a PICNN, i.e., $\mathcal{T}_{\theta} = \nabla \psi_\theta^* = \nabla_1 \text{PICNN}_\theta$.
The PICNN is then trained using the entropy-regularized Wasserstein distance \eqref{eq:reg-ot} between the predictions $\hat{\nu} = \nabla \psi^* _\sharp \mu = \nabla_1 \text{PICNN}_\theta (\cdot, c)_\sharp \mu$ given source samples $\mu$ and condition $c$ and the observed target population $\nu$ as a loss function, i.e.,
\begin{equation} \label{eq:pot_loss}
    \ell_\text{POT}(\mu, \nu, c; \theta) = \We(\nabla_1 \text{PICNN}_\theta (\cdot, c)_\sharp\mu,\nu).
\end{equation}

Throughout this work, we choose a different route and propose instead to learn $\mathcal{T}_{\theta}$ via the dual optimal transport problem. We consider the strategy proposed by \citet{makkuva2020optimal} and utilized by \citet{bunne2021learning} in the context of single-cell perturbation analyses.
 $\mathcal{T}_{\theta}$ is then parameterized via the pair of dual potentials $f$ and $g$, which  themselves are defined by a pair of PICNNs $g:\text{PICNN}_{\theta_g}(\cdot,c)$ and $f:\text{PICNN}_{\theta_f}(\cdot,c)$ such that $\hat{\nu} = \nabla g _\sharp \mu = \nabla_1 \text{PICNN}_{\theta_g} (\cdot,c)\sharp \mu$ is approximately $\nu$, as well as $\hat{\mu} = \nabla f _\sharp \nu = \nabla_1 \text{PICNN}_{\theta_f} (\cdot, c)_\sharp\nu$ is approximately $\mu$ on a labeled observation $((\mu,\nu),c)$ with parameters $\theta = (\theta_g, \theta_f)$.
In order to optimize the pair of PICNNs, which parameterize the two dual functions, \citet{makkuva2020optimal} derive an approximate formulation of \eqref{eq:dual}.
First, \citet[Theorem 2.9]{villani2021topics} rephrases~\eqref{eq:dual} over the pair of dual potentials $(f, g)$ to
\begin{equation} \label{eq:dual-ot-cvx}
    W(\mu, \nu)= \underbrace{\frac{1}{2}\mathbb{E}\left[\|x\|_{2}^{2}+\|y\|_{2}^{2}\right]}_{\mathcal{C}_{\mu, \nu}}-\inf _{f \text{ convex}} \mathbb{E}_{\mu}[f(X)]+\mathbb{E}_{\nu}\left[f^{*}(Y)\right],
\end{equation}
where $f^*(y) = \sup_x \dotp{x}{y} - f(x)$ is $f$'s convex conjugate.
In a second step, \citet{makkuva2020optimal} derive a min-max formulation by approximating the convex conjugate in \eqref{eq:dual-ot-cvx} via
\begin{equation} \label{eq:ot-minmax}
    W(\mu, \nu)=\sup _{\substack{f  \text{ convex} \\ f^{*} \in L^{1}(\nu)}} \newinf _{g \text{ convex}}  \mathcal{C}_{\mu, \nu} - \underbrace{\mathbb{E}_{\mu}[f(x)]-\mathbb{E}_{\nu}[\langle y, \nabla g(y)\rangle-f(\nabla g(y))]}_{\mathcal{V}_{\mu, \nu}(f, g)},
\end{equation}
and by relaxing the constraints on $g$. 
Thus, the dual potentials $f$ and $g$ can be learned via an alternate min-max optimization problem with loss functions
\begin{align} 
    \ell_\text{DOT}^f(\mu, \nu, c; \theta_f) &= \mathbb{E}_{x \sim \mu}[\text{PICNN}_{\theta_g}(x, c)] - \mathbb{E}_{y \sim \nu}[\text{PICNN}_{\theta_f}(\nabla \text{PICNN}_{\theta_g}(y, c), c)], \text{ and }  \label{eq:makkuva_f_loss} \\
    \ell_\text{DOT}^g(\mu, \nu, c; \theta_g) &= -\mathbb{E}_{y \sim \nu}[\langle y, \nabla \text{PICNN}_{\theta_g}(y, c)\rangle-\text{PICNN}_{\theta_f}(\nabla \text{PICNN}_{\theta_g}(y, c), c)]. \label{eq:makkuva_g_loss}
\end{align}
For more details, see \citet{makkuva2020optimal, korotin2021neural}.

Thus, dependent on the strategy chosen, $\mathcal{T}_{\theta}$ is parameterized via a single or a pair of PICNN. Each network takes as input the source distribution $\mu$ ---in which it is input convex--- as well as an embedded context variable $\hat{c}$, returned by combinator $\mathcal{C}_\Phi$ and embedding module $\mathcal{E}_\phi$.
Parameters of all three modules are jointly trained based on the derived optimal transport loss $\ell = \{\ell_\text{POT}, \ell_\text{DOT}\}$, which measures how close predicted target cells $\hat{\nu}$ are from the observed target population $\nu$, given source population $\mu$ and context $c$ as inputs.

\SetKwComment{Comment}{\# }{}
\begin{algorithm}[t]
\KwIn{Dataset $\mathcal{D}=\{\mu_i, \nu_i, c_i \}_{i=0}^N$ of $N$ pairs of populations before $\mu_i$ and after transport $\nu_i$ connected to a context $c_i$, $\theta^0$ transport map $\mathcal{T}$ parameter initialization, $\phi^0$ embedding $\mathcal{E}$ parameter initialization, $\Phi^0$ embedding $\mathcal{C}$ parameter initialization, learning rates $\text{lr}_\theta$, $\text{lr}_\phi$ and $\text{lr}_\Phi$, and flag which \texttt{loss} function to use.
In the case of the dual, we have $\theta = (\theta_f, \theta_g)$ parameterizing the dual potentials $f$ and $g$ and \texttt{train\_freq\_f} specifies the training frequency of dual potential $f$.}
\KwOut{Transport map $\mathcal{T}_\theta$, embedding $\mathcal{E}_\phi$, and combinator $\mathcal{C}_\Phi$.}

$\theta, \phi, \Phi \leftarrow \theta^0, \phi^0, \Phi^0$

\For{$\{\mu_i, \nu_i, c_i \} \in \mathcal{D}$} {

\Comment{Split (combination) context $c_i$ into individual contexts.} 

  $c_i^1, c_i^2, \dots, c_i^k = c_i\,\,$ 

  $\hat{c}_i = \mathcal{C}_\Phi(\mathcal{E}_\phi(c_i^1), \mathcal{E}_\phi(c_i^2), \dots, \mathcal{E}_\phi(c_i^k))$

  \If{\texttt{setting} $==$ `dual'}{
      \If{i \% \texttt{train\_freq\_f} $==$ 0}{

        $\ell \leftarrow \ell_\text{DOT}^f(\mu_i, \nu_i, \hat{c}_i; \theta_f)\enspace$ \eqref{eq:makkuva_f_loss}
      }
      \Else{$\ell \leftarrow \ell_\text{DOT}^g(\mu_i, \nu_i, \hat{c}_i; \theta_g)\enspace$  \eqref{eq:makkuva_g_loss}}
}
  \Else{$\ell \leftarrow \ell_\text{POT}(\mu_i, \nu_i, \hat{c}_i; \theta)\enspace$ \eqref{eq:pot_loss}}
  
  \Comment{Jointly optimize parameters $\theta, \phi, \Phi$ given loss $\ell$.} 

  $\theta \leftarrow \theta - \text{lr}_\theta \times \nabla_\theta \ell$

  $\phi \leftarrow \phi - \text{lr}_\phi \times \nabla_\phi\ell$

  $\Phi \leftarrow \Phi - \text{lr}_\Phi \times \nabla_\Phi\ell$
}
\Return{}
\caption{\textsc{CondOT} Algorithm.}
\label{algo:condot}
\end{algorithm}

\clearpage
\newpage
\section{Additional Experimental Results} \label{app:add_exps}

\newtext{\subsection{Comparison of Initialization Methods} \label{app:comp_init}

\begin{figure}[H]
    \centering
    \includegraphics[width=1\textwidth]{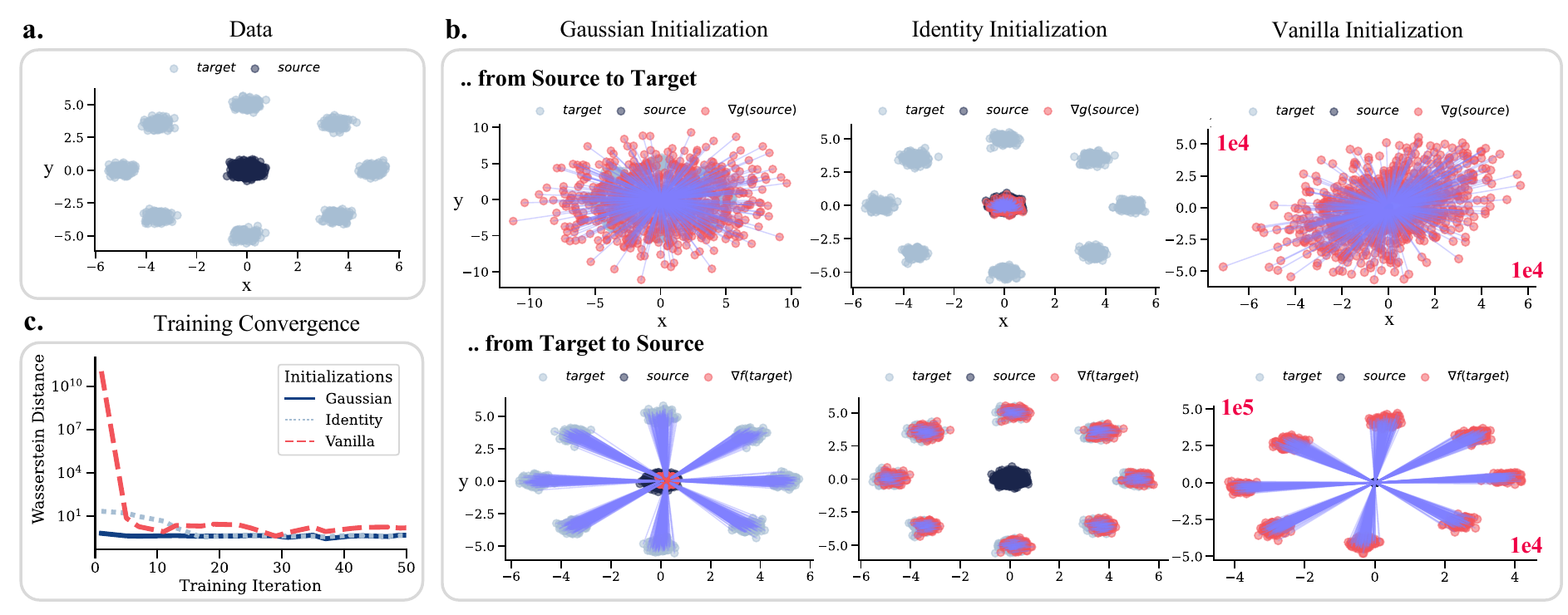}
    \caption{\looseness -1 \newtext{Comparison of ICNN initialization methods on a \textbf{a.} synthetic dataset containing source (dark blue) and target samples (light blue). \textbf{b.} Predicted samples (red) are obtained by transporting source samples (with dual potential $g$, first row) or target samples (with dual potential $f$, second row) to match the respective observations. The ICNNs are initialized such that they resemble a Gaussian closed-form approximation, the identity, or a random vanilla map (more details in \S~\ref{sec:inits}). Without any pretraining, the Gaussian initialization transports the samples to the Gaussian approximation of the respective target distribution. The identity initialization mimics the identity map and thus does not move the samples. The na\"ive vanilla initialization, on the other hand, starts with a solution far off from the target (i.e., values are in the range of $1e4$ or $1e5$). \textbf{c.} The chosen initialization strongly affects the convergence of the solution over the course of the training, here measured by the Wasserstein distance.}}
    \label{fig:exp_comp_inits}
\end{figure}

\looseness -1 We conduct a simple experiment based on a synthetic dataset displayed in Fig.~\ref{fig:exp_comp_inits}a, in which we seek to learn a mapping between source and target samples by parameterizing the dual potentials $f$ and $g$ with two ICNNs based on different initialization schemes (see Algorithm~\ref{algo:condot}, \S~\ref{app:algorithm}).
To showcase different initialization methods, we compare the initial predictions (at training iteration 0) of transported samples for the vanilla, the identity, and the Gaussian initialization (Fig.~\ref{fig:exp_comp_inits}b).
As the Gaussian initialization instantiates maps which transport source samples to the Gaussian approximation of the target samples, the initialization already captures well the source or target distribution using ICNN $f$ or $g$, respectively (Fig.~\ref{fig:exp_comp_inits}b, first column).
The identity initialization, instead, configures maps which do not move the samples from the initial distribution (see Fig.~\ref{fig:exp_comp_inits}b, second column).
Both initialization schemes proposed in this work thus result in map parameterizations, which initially (before training) realize non-trivial and admissible Monge maps. The vanilla initialization, on the other hand, instantiates random maps, mapping the point far away from the source and target distribution, thus impeding fast and robust training (see Fig.~\ref{fig:exp_comp_inits}, third column).
We want to stress that this is achieved without costly and elaborate pretraining of the networks as proposed in \citet[Appendix B]{korotin2020wasserstein}.

\looseness -1 The selected initialization method also strongly affects the convergence of the solution over the course of the training, which we monitor using the Wasserstein distance between observed and predicted target samples (see Fig.~\ref{fig:exp_comp_inits}c).
In this simple example, the Gaussian initialization is already close to the solution, thus the Wasserstein distance resembles the final solution already at the beginning of the training.
The identity initialization similarly starts with a mapping closer to the solution compared to a random vanilla initialization, thus achieving fast and stable convergence of the min-max optimization problem \eqref{eq:ot-minmax}. This experiment thus demonstrates that a proper initialization is not only crucial for fast convergence but also the overall robustness of training and result.}

\clearpage
\newpage
\subsection{Out-of-Sample Predictions in Unknown Contexts}

\begin{figure}[H]
    \centering
    \includegraphics[width=1.2\textwidth]{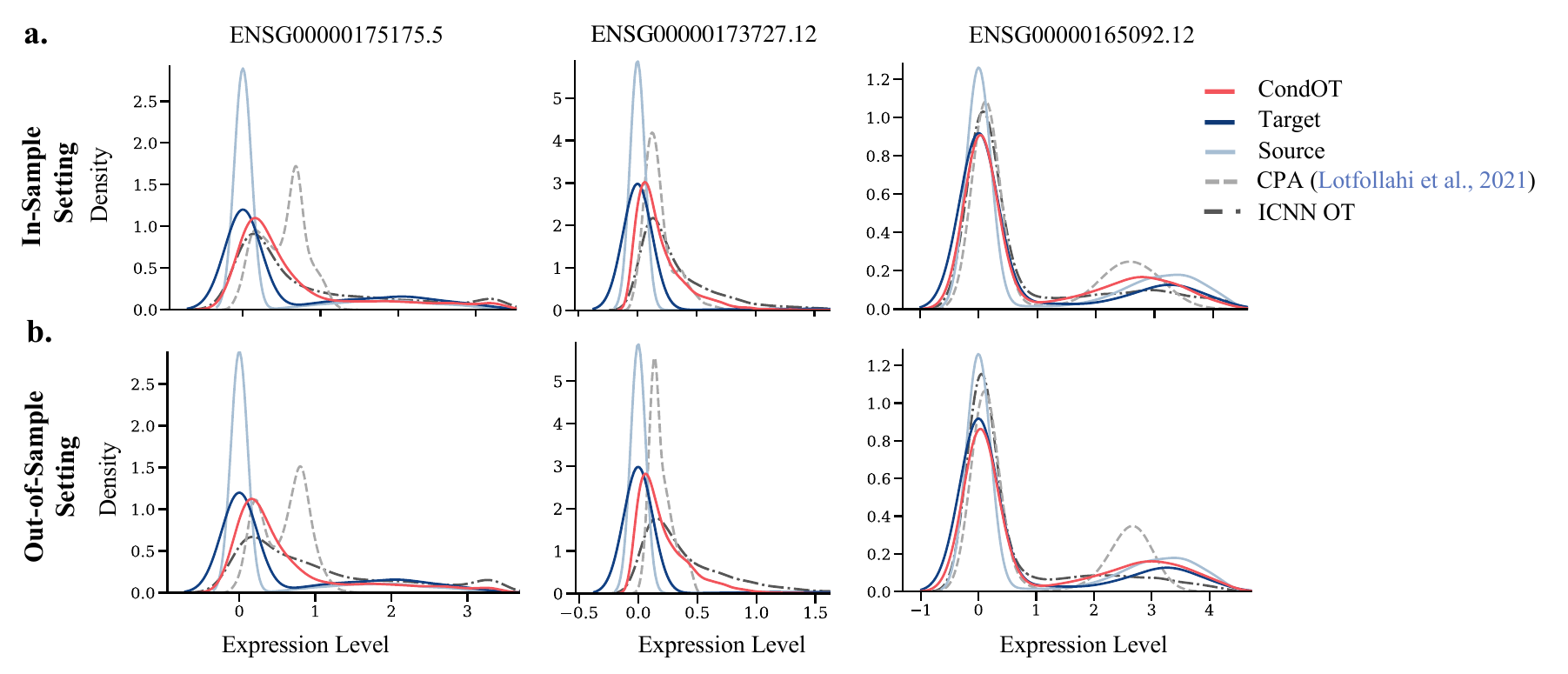}
    \caption{Marginal distributions of observed source (light blue) and target distributions (dark blue), as well as predictions on perturbed distributions by \textsc{CondOT} (red) and baselines (gray) of different genes \textbf{a.} in the in-sample setting, where dosage $100$nM was seen during training, and \textbf{b.} out-of-sample setting, where dosage $100$nM was \emph{not} seen during training. Predicted marginals of each method should match the marginal of the target population (dark blue. While the performance of \textsc{CondOT} is consistent from the in-sample to the out-of-sample setting, both baselines show differences. These differences are subtle, however, only $3$ out of $1,000$ genes are displayed.}
    \label{fig:exp_scalar_sciplex_marginals}
\end{figure}

\begin{figure}[H]
    \centering
    \includegraphics[width=1.1\textwidth]{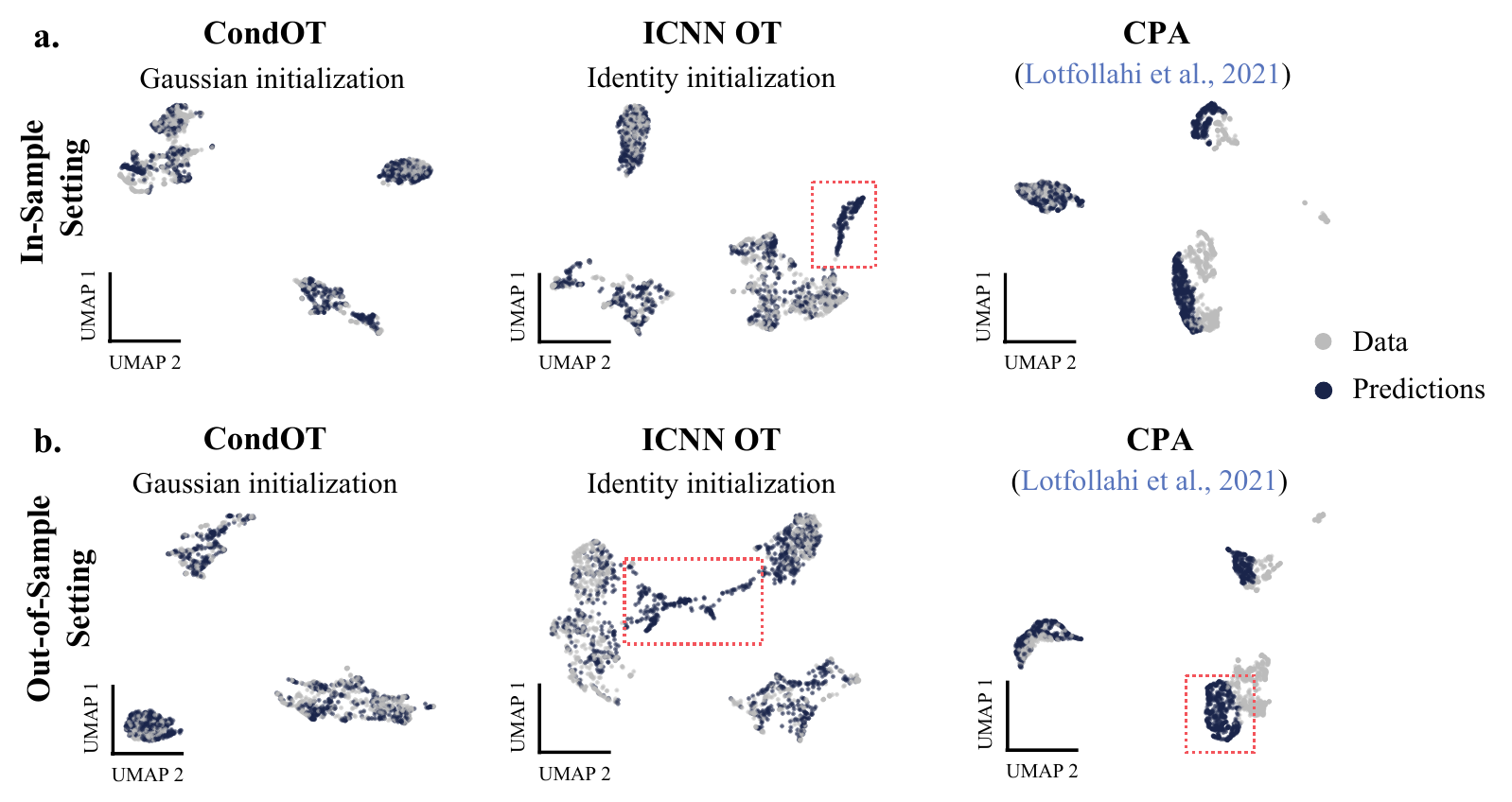}
    \caption{UMAP embeddings of cells perturbed by Givinostat dosage $100$nM (gray) and predictions of \textsc{CondOT} (ours), \textsc{ICNN OT}~\citep{makkuva2020optimal}, and \textsc{CPA}~\citep{lotfollahi2021compositional} (blue). Contrary to the out-of-sample setting, the dosage $100$nM was seen during training in the in-sample setting. While \textsc{CondOT} covers the space of observed perturbed cells, the baselines fail to capture subpopulations (see red squares).}
    \label{fig:exp_scalar_sciplex_umaps}
\end{figure}

\subsection{Predicting Unknown Perturbations and Perturbations in Combination}

\begin{figure}[H]
    \centering
    \includegraphics[width=1\textwidth]{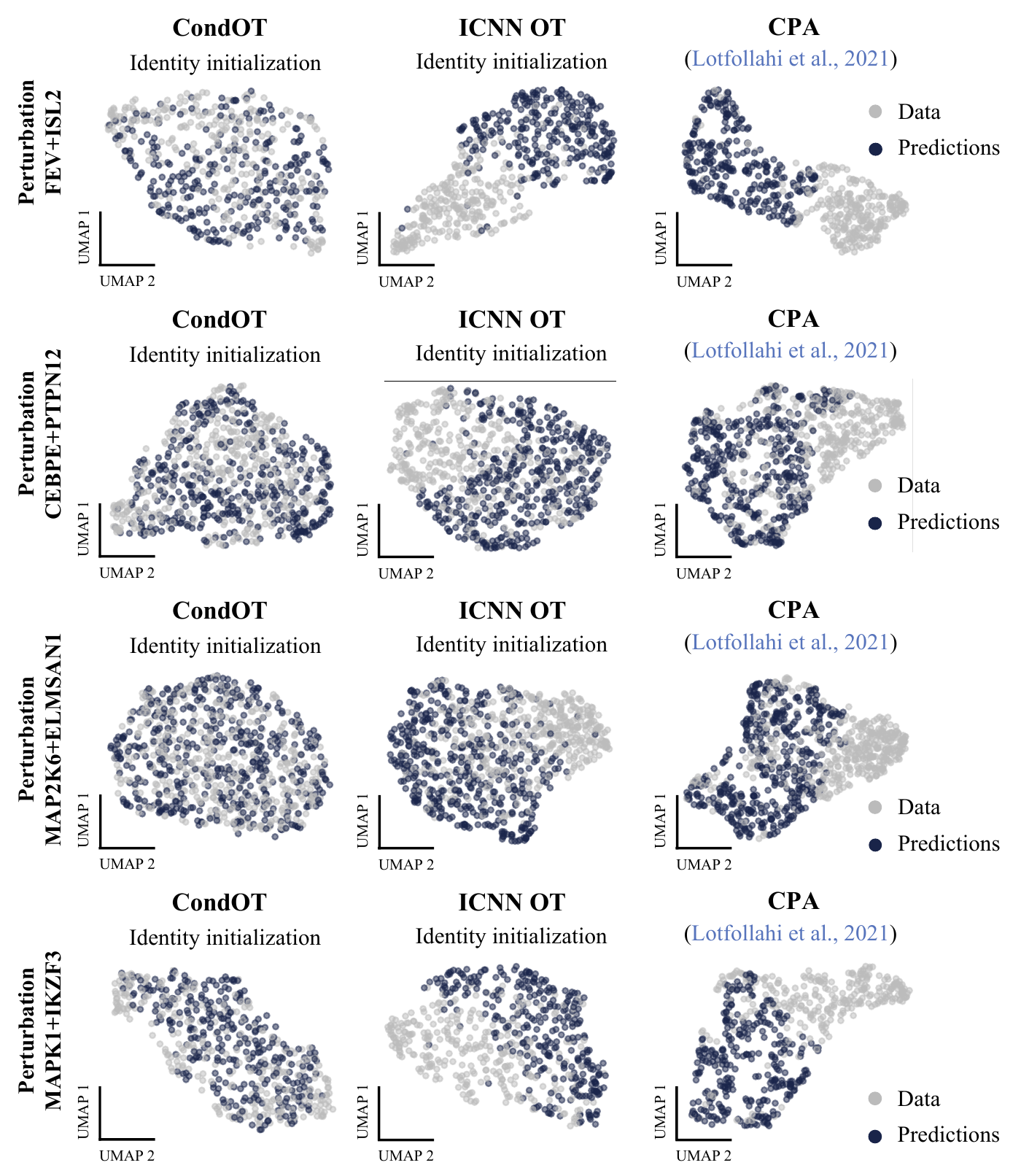}
    \caption{UMAP embeddings of cells perturbed by different combinations (grey) and predictions of \textsc{CondOT} (ours), \textsc{ICNN OT}~\citep{makkuva2020optimal}, and \textsc{CPA}~\citep{lotfollahi2021compositional} (blue). While \textsc{CondOT} covers the space of observed perturbed cells, the baselines fail to capture subpopulations.}
    \label{fig:exp_scalar_norman_umaps_more}
\end{figure}

\subsection{Comparing Multi-Task Performance to Ideal Single-Task Baseline} \label{app:add_baseline}

\begin{table}[h]
\caption{Wasserstein loss $\We$ of different methods on the top-50 marker genes for the drugs Givinostat and Trametinib, where we conduct the analysis for different dosages individually on the dataset by \citet{srivatsan2020massively}.
}
\label{tab:lower_bound}
\adjustbox{max width=\linewidth}{%
\begin{tabular}{lcccc}
\toprule
 & \multicolumn{4}{c}{\textbf{Wasserstein Loss} $\We$} \\ \cline{2-5} 
\diagbox{\textbf{Model}}{\textbf{Dosages}} & $10\,$nM & $100\,$nM & $1,000\,$nM & $10,000\,$nM \\ \midrule
\textsc{CPA} \citep{lotfollahi2021compositional} & 13.75 $\pm$ 1.41 & 13.75 $\pm$ 0.93 & 15.81 $\pm$ 2.16 & 55.12 $\pm$ 58.14 \\
\textsc{ICNN OT} \citep{makkuva2020optimal} (on all conditions) & 12.37 $\pm$ 1.66 & 12.53 $\pm$ 2.40 & 13.36 $\pm$ 3.02 & 31.02 $\pm$ 28.12 \\
\textsc{ICNN OT} \citep{makkuva2020optimal} (on selected condition) & 10.98 $\pm$ 1.15 & 10.37 $\pm$ 0.23 & 10.58 $\pm$ 1.93 & 20.55 $\pm$ 14.16 \\
\textsc{CondOT} (Identity initialization) & 10.54 $\pm$ 0.37 & 10.58 $\pm$ 0.04 & 12.13 $\pm$ 2.43 & 20.97 $\pm$ 15.02 \\
\textsc{CondOT} (Gaussian initialization) & 10.56 $\pm$ 0.45 & 10.54 $\pm$ 0.16 & 12.12 $\pm$ 2.26 & 21.30 $\pm$ 15.29 \\ \bottomrule
\end{tabular}}
\end{table}

In \S~\ref{sec:eval_scalar}~and~~\ref{sec:eval_covariate} (Table~\ref{tab:exp_scalar_covariate_sciplex}), we compared \textsc{CondOT} with different initializations against current state-of-the-art methods \citep[\textsc{CPA}]{lotfollahi2021compositional} and previous neural optimal transport-based methods used on single-cell data \citep[\textsc{ICNN OT}]{makkuva2020optimal}. To challenge the performance of \textsc{CondOT} even further, we add another baseline in which we train \textsc{ICNN OT} individually for each distinct condition. This baseline can be seen as a \textit{lower bound} on what accuracy \textsc{CondOT} can reach based on modeling perturbation responses by parameterizing Monge maps.
In particular, we train \textsc{CondOT} and both baselines (\textsc{ICNN OT} and \textsc{CPA}) to predict the dosage-dependent perturbation response to two different drugs, Trametinib and Givinostat. While both \textsc{CondOT} and \textsc{CPA} allow us to condition the training on all respective dosages, we train two variants of \textsc{ICNN OT}: The first version is trained on all conditions, while the additional baseline (the lower bound, i.e., \textsc{ICNN OT} selected condition) computes different and independent \textsc{ICNN OT} models for each dosage. While this would fail to generalize to new contexts and it requires all contexts to be distinctly known, this is, in a way, the best we can expect to achieve.
We believe the setting in which we condition on scalars is a good start because in this 1D setting for $c$, the inability to generalize is less critical (as opposed to predicting previously unobserved  combinations of drugs).

The results on on 50 marker genes in data space with 1000 genes are displayed in the Table~\ref{tab:lower_bound}. This additional experiment clearly demonstrates that \textsc{CondOT} predicts perturbation responses as well as a baseline which was trained purely on individual conditions, while still being able to generalize (see Table~\ref{tab:exp_scalar_covariate_sciplex}). As often mentioned in the multitask learning literature \citep{mahabadi2021parameter}, sharing of parameters (the PICNN) and conditioning seems to improve by increasing the effective sample size of the problem.

\section{Datasets} \label{app:datasets}

We evaluate \textsc{CondOT} on different tasks, consisting of a pair of source $\mu$ and target measures $\nu$, as well as context variables $c$ of different nature.
In particular, we consider single-cell datasets in which populations of single cells have been monitored with modern high-throughput methods such as single-cell RNA sequencing technologies.
Characterizing and modeling perturbation responses at the level of single cells with access to \emph{unpaired} populations of control and perturbed cells remains one of the grand challenges of biology.
In this work, we consider the task of modeling molecular responses to cancer drugs with context variables being the drug's dosage (i.e., a scalar, \S~\ref{sec:eval_scalar}) as well as covariates such as different cancer cell lines present in the population (\S~\ref{sec:eval_scalar}).
Further, we study cellular responses to genetic perturbations, where we condition on the perturbation, i.e., action, chosen. Here we differentiate between settings where we encounter \emph{known} actions (\S~\ref{sec:eval_action_known}), \emph{unknown} actions (\S~\ref{sec:eval_action_unknown}), and action applied in combination (\S~\ref{sec:eval_action_comb}) during evaluation.
In the following, we introduce the datasets in more depth, describe preprocessing steps, feature selection, and data splits.

\subsection{... by \citet{srivatsan2020massively}}
\looseness -1 Cancer drugs reduce uncontrolled cell growth and proliferation by inhibiting DNA replication and RNA transcription as well as targeting proteins crucial for cancer progression. In doing so, they modulate downstream signaling cascades, affect cell growth and morphology, and alter gene expression profiles of single cells. 
\citet{srivatsan2020massively} conduct a scRNA-seq–based phenotyping screen of transcriptional responses to thousands of independent perturbations at single-cell resolution.
The measured cell population contains three well-characterized cancer cell lines, including A549, a human lung adenocarcinoma, K562, a chronic myelogenous leukemia, and MCF7, a mammary adenocarcinoma cell line.
Due to different transcriptional profiles of each cancer cell line, drug compounds might cause divergent cellular responses in each subpopulation.
For our analysis, we consider the drug Givinostat, a histone deacetylase inhibitor with potential anti-inflammatory, anti-angiogenic, and antineoplastic activities \citep{rambaldi2010pilot}.
The dataset contains $17,565$ control cells as well as $3,541$ cells perturbed by Givinostat with different dosages, i.e., $10\,$nM, $100\,$nM, $1,000\,$nM, $10,000\,$nM.

\paragraph{Data Preprocessing}
The data is available for download in the Gene Expression Omnibus (GEO) database under accession number \href{https://www.ncbi.nlm.nih.gov/geo/query/acc.cgi?acc=GSM4150378}{GSM4150378}.
For data quality control and preprocessing, we follow the analysis of \citet{lotfollahi2021compositional}. The count matrix obtained from GEO consists of $581,777$ cells. The data was subset to half its size, with $290,888$ cells remaining after quality control for all $188$ different compounds. We proceeded with log-transformation and the the selection of $1,000$ highly-variant genes using \texttt{scanpy} \citep{wolf2018scanpy}.

\paragraph{Feature Selection}
Single-cell RNA sequencing data is very high-dimensional, even after selecting $1,000$ highly-variant genes.
For the downstream analysis of how well the overall perturbation effect has been captured, we thus select the top $50$ marker genes, i.e., those genes which show strong differences between perturbed and unperturbed states. This analysis is conducted based on the \texttt{scanpy}'s function \texttt{rank\_genes\_groups}, setting unperturbed cells as reference ~\citep{wolf2018scanpy}. It is important to note here, that \textsc{CondOT} operates on the full dataset and the marker genes are only considered to report meaningful evaluation measures.

\begin{figure}
    \centering
    \includegraphics[width=\textwidth]{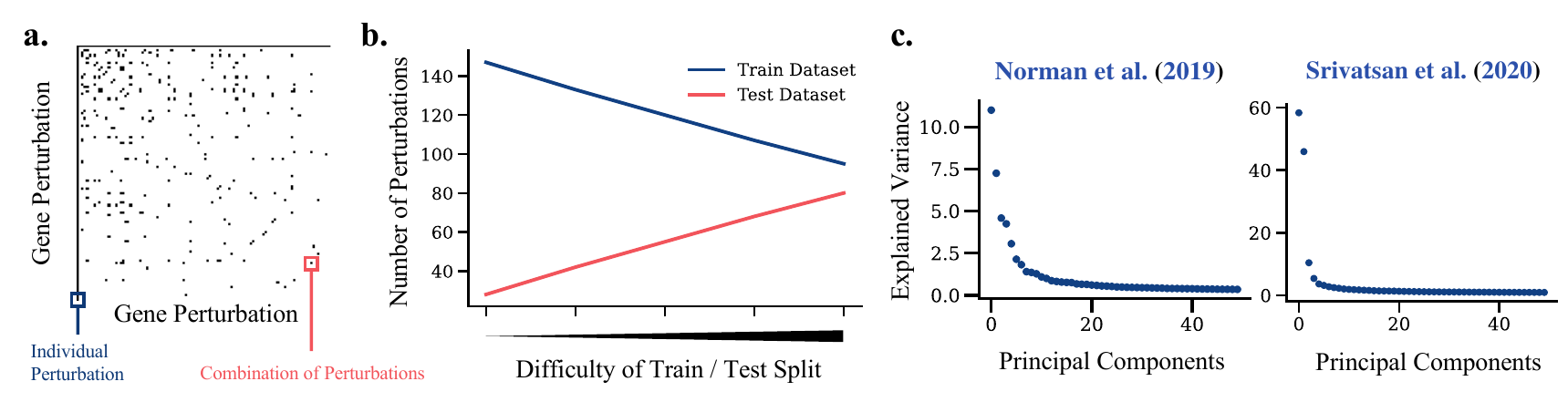}
    \caption{\textbf{a.} The indicator matrix of all individual perturbations as well as those perturbation pairs available in combination (black) in the dataset by \citet{norman2019exploring}. \textbf{b.} Size of the different train / test splits of the dataset by \citet{norman2019exploring}. The train set contains all single perturbations as well as a decreasing number of combinations with increasing difficulty of the data split. For more details, see~\S\ref{app:datasplits}. \textbf{c.} Explained variance of different datasets per principal component.}
    \label{fig:datasets_processing}
\end{figure}

\subsection{... by \citet{norman2019exploring}} 
\looseness -1 Genetic interactions and their joint expression give rise to an inconceivable organismal complexity and uncountable many diverse phenotypes and behaviors.
Constructing a systematic genetic interaction map is crucial for a better understanding of cellular mechanisms in health and disease.
Thus, \citet{norman2019exploring} conducted single-cell, pooled transcriptional profiling of CRISPR-mediated perturbations to link genetic perturbation to its transcriptional consequences using the Perturb-Seq technology \citep{dixit2016perturb}.
The dataset consists of individual perturbations as well as joint knockouts of different genes, allowing us to study the phenotypic consequences of perturbing a pair of genes alone or in combination. The indicator matrix of all individual perturbations as well as those pairs available in combination can be found in Fig.~\ref{fig:datasets_processing}a.

\paragraph{Data Preprocessing}
\looseness -1 The data is available for download in the Gene Expression Omnibus (GEO) database under accession number \href{https://www.ncbi.nlm.nih.gov/geo/query/acc.cgi?acc=GSE133344}{GSE133344}.
For data quality control and preprocessing, we follow the analysis of \citet{lotfollahi2021compositional}. 
We discarded those genetic perturbations with less than $250$ cells, resulting in a dataset with $92$ individual perturbations and $84$ perturbations in combination.
This further included, the exclusion of particular subsets of control cells with in total $98,419$ remaining, data normalization, log-transformation, and selection of $1,500$ highly-variant genes using \texttt{scanpy} \citep{wolf2018scanpy}.

\paragraph{Feature Selection}
Similar as above, for evaluation we select the top $50$ marker genes, i.e., those genes most strongly affected by the particular genetic perturbation.

\paragraph{Data Splits} \label{app:datasplits}
Following \citet{lotfollahi2021compositional}, we create different train / test dataset splits of increasing difficulty.
The train splits hereby always contain all $92$ individual perturbations as well as varying numbers of combinations. The easiest train split contains 55 perturbations, while the test set only carries 28 combinations which are unknown in the evaluation. Consecutive splits get increasingly harder, comprising 42, 29, 16, and 4 combinations in the train set (besides all single perturbations) and 41, 54, 67, and 79 combinations in the test set, respectively (see Fig.~\ref{fig:datasets_processing}b).

\section{Experimental Details} \label{app:exp_details}
In the following, we describe the experimental setup by providing an overview on the baselines, evaluation metrics, parameterizations of \textsc{CondOT}'s modules, and hyperparameters chosen.

\subsection{Baselines} \label{app:baselines}
We consider several baselines to put \textsc{CondOT}'s performance into perspective. This includes current state-of-the-art methods, as well as ablations of our methods.

\paragraph{Compositional Perturbation Autoencoder (\textsc{CPA})} Building up on previous work \citep{lotfollahi2019scgen, lotfollahi2020conditional}, the current state-of-the-art approach conditional perturbation autoencoder (\textsc{CPA}) learns transcriptional perturbation responses across different cell types, applied dosages, and perturbation combinations \citep{lotfollahi2021compositional}.
The architecture hereby consists of several modules. \textsc{CPA} predicts perturbed states of populations by learning a factorized latent representation of both perturbations and covariates, with separate embeddings for particle feature vectors, perturbations, and external covariates. These embeddings are independent of each other by design to later allow modular recombination of different modules and thus allowing the model to make predictions on unseen perturbations in combination.
We follow the experimental setup outlined in \citep{lotfollahi2021compositional}. Similarly to perturbations, covariates such as cell type or dosage are encoded via one-hot vectors. Thus, \textsc{CPA} can not be utilized to make predictions on \textit{unknown} perturbations as studied in \S~\ref{sec:eval_action_unknown}.

\paragraph{\textsc{ICNN OT}} A crucial ablation study of \textsc{CondOT} is to learn the transition of source population $\mu$ to target population $\nu$ \emph{without} considering context $c$. Thus, we use standard ICNNs \eqref{eq:icnn} to parameterize the transport map module $\mathcal{T}_\theta$ via two dual potentials as proposed in \citet{makkuva2020optimal} and \citet{bunne2021learning}. As for the PICNNs, we utilize different initialization schemes as derived in \S~\ref{sec:inits}. 

\subsection{Evaluation Metrics} \label{app:eval_metrics}
\looseness -1 Since we lack access to the ground truth pair of perturbed and unperturbed observations on the single cell level, we consider evaluation metrics on the level of the distribution of real and predicted perturbation states to analyze the effectiveness of \textsc{CondOT}.
We report results based on several metrics:

\paragraph{Wasserstein Distance} We measure accuracy of the predicted target population $\hat{\nu}$ to the observed target population $\nu$ using the entropy-regularized Wasserstein distance \citep{cuturi2013sinkhorn} provided in the \texttt{OTT} library \citep{cuturi2022optimal} defined as
\begin{equation}\label{eq:reg-ot}
\We(\hat{\nu},\nu) \defeq \min_{\bP\in U(\hat{\nu},\nu)} \dotp{\bP}{[\|x_i - y_j\|^2]_{ij}}  \,-\varepsilon H(\bP),
\end{equation}
where $H(\bP) \defeq -\sum_{ij} \bP_{ij} (\log \bP_{ij} - 1)$ and the polytope $U(\hat{\nu},\nu)$ is the set of $n\times m$ matrices $\{\bP\in\mathbb{R}^{n \times m}_+, \bP\mathbf{1}_m = \hat{\nu}, \bP^\top\mathbf{1}_n=\nu\}$.
Throughout the evaluation, we set $\varepsilon = 0.1$.

\paragraph{Maximum Mean Discrepancy} Kernel maximum mean discrepancy~\citep{gretton2012kernel} is another metric to measure distances between distributions, i.e., for our purpose between the predicted target population $\hat{\nu}$ to the observed target population $\nu$.
Given two random variables $x$ and $y$ with distributions $\hat{\nu}$ and $\nu$, and a kernel function $\omega$, \citet{gretton2012kernel} define the squared MMD as:
\begin{equation*}
    \text{MMD}(\hat{\nu},\nu; \omega) = \mathbb{E}_{x,x^\prime}[\omega(x, x^\prime)] + \mathbb{E}_{y,y^\prime}[\omega(y, y^\prime)] - 2\mathbb{E}_{x,y}[\omega(x, y)].
\end{equation*}
We report an unbiased estimate of $\text{MMD}(\hat{\nu},\nu)$, in which the expectations are evaluated by averages over the population particles in each set. We utilize the RBF kernel, and as is usually done, report the MMD as an average over several length scales, i.e., $2, 1, 0.5, 0.1, 0.01, 0.005$.

\paragraph{Perturbation Signatures}
A common method to quantify the effect of a perturbation on a population is to compute its perturbation signature \citep[(PS)]{stathias2018drug}, computed via the difference in means between the distribution of perturbed states and control states of each feature, e.g., here individual genes. $\ell_2$(PS) then refers to the $\ell_2$-distance between the perturbation signatures computed on the observed and predicted distributions, $\nu$ and $\hat{\nu}$. As before, let $\mu$ be the set of observed unperturbed population particles, $\nu$ the set of observed perturbed particles, as well as $\hat{\nu}$ the predicted perturbed state of population $\mu$. The $\ell_2$(PS) is then defined as
\begin{equation*}
    \text{PS}(\nu, \mu) = \frac{1}{m}\sum_{y_i \in \nu}{y_i} - \frac{1}{n}\sum_{x_i \in \mu}{x_i},
\end{equation*}
where $n$ is the size of the unperturbed and $m$ of the perturbed population.
We report the $\ell_2$ distance between the observed signature $\text{PS}(\nu, \mu)$ and the predicted signature $\text{PS}(\hat{\nu}, \mu)$, which is equivalent to simply computing the difference in the means between the observed and predicted distributions.

\subsection{\textsc{CondOT} Modules} \label{app:modules}

\textsc{CondOT} consists of several modules for which different choices can be considered. Here, we provide a brief overview on the options and their parameterization.

\subsubsection{Embedding Module $\mathcal{E}$}
The embedding module allows us to consider context variables $c$ of various nature. In the case of scalars, no sophisticated embedding is necessary. In contrast, covariate contexts as well as potentially complex action descriptions require embeddings in order to be processed by the combinator $\mathcal{C}$, and transport map module $\mathcal{T}$.

\paragraph{One-Hot Embedding $\mathcal{E}_\text{ohe}$}
Covariates, such as subpopulation or patient identifiers, can be simply embedded via one-hot encodings. These embeddings, however, are not able to capture unknown covariates after training.

\paragraph{Mode-of-Action Embedding $\mathcal{E}_\text{moa}$}
In certain cases, actions might possess distinct properties which allow for a direct embeddings using this domain knowledge, i.e., molecular representations for molecules \citep{rong2020grover, rogers2010extended}.
In the case of genetic perturbations, however, no straightforward embedding is available.
We thus introduce so-called mode-of-action embeddings, which map actions into a latent space-based on their mechanism of action and effect on the target population.
In the fashion of word embeddings \citep{mikolov2013efficient, mikolov2013distributed, mikolov2013linguistic}, we require actions with similar effect to be closely embedded in the learned representation.
This means, however, that we require some sample access of target population particles, i.e., perturbed cells by individual compounds (not in combination).
While several metric embeddings are possible \citep{chopra2005learning}, we here test a simple multi-dimensional scaling-based embedding \citep{mead1992review}.
For this we compute the pairwise Wasserstein distance matrix between all target populations of different individual perturbations. We then compute a 10-dimensional MDS embedding-based on the stress minimization using majorization algorithm (smacof) \citep{de2009multidimensional} of \texttt{sklearn} \citep{pedregosa2011scikit}, which serves as descriptor for each individual perturbation.

\subsubsection{Combinator Module $\mathcal{C}$}
The combinator module allows us to pass an arbitrary number of context $c$ to the transport map module $\mathcal{T}$.

\paragraph{Multi-Hot Combinator $\mathcal{C}^\text{ohe}_+$}
A na\"ive way of constructing the combinator is to combine different actions via a multi-hot encoding. If all single perturbations are observed during training, each individual action can be represented via a one-hot encoding. The potential combination of different actions, is then encoded by adding the respective one-hot encodings, resulting in a multi-hot encoding for each combination.
A limitation of this embedding, however, is that it cannot generalize to unknown action after training.

\paragraph{Deep Set Combinator $\mathcal{C}^\text{moa}_\Phi$}
When not considering one-hot-based embeddings and when aiming to generalize to unseen perturbations, we need a combinator module which learns how to associate different individual embeddings with each other to receive a joint embedding.
As we for now do not make an assumption on the order of the perturbation, we consider a permutation-invariance network architecture such as deep sets~\citep{zaheer2017} with parameters $\Phi$. Taking a set of arbitrary size $k$ containing individual context embeddings $\{\mathcal{E}_\text{moa}(c^1), \mathcal{E}_\text{moa}(c^2), \dots, \mathcal{E}_\text{moa}(c^k)\}$, it returns a learned combination embedding $\hat{c}_i = \mathcal{C}_\Phi(\mathcal{E}_\text{moa}(c^1), \mathcal{E}_\text{moa}(c^2), \dots, \mathcal{E}_\text{moa}(c^k))$.

\subsubsection{Transport Map Module $\mathcal{T}$}
The transport map module takes as input samples of the source distribution $\mu$ as well as context $c$ and returns the perturbed population $\nu$.
Map $\mathcal{T}_\theta$ is thereby parameterized via PICNNs as we require input convexity in $\mu$ but not $c$. In the case where we consider learning $\mathcal{T}_\theta$ via the dual \eqref{eq:dual}, it is defined by a pair of PICNNs with parameters $\theta = (\theta_f, \theta_g)$, parameterizing the set of dual variables $f$ and $g$. When deploying the primal OT problem \eqref{eq:monge}, we parameterize a single Brenier potential via a PICNN with parameters $\theta$.

As suggested by \citet{makkuva2020optimal}, we relax the convexity constraint on PICNN $g$ and instead penalize its negative weights $W^z_k$
\begin{equation*}
    R\left(\theta\right)=\lambda \sum_{W^z_k \in \theta}\left\|\max \left(-W^z_k, 0\right)\right\|_{F}^{2}.
\end{equation*}
The convexity constraint on PICNN $f$ is enforced after each update by setting the negative weights of all $W^z_k \in \theta_f $ to zero.
Thus, the full objective then states
\begin{align*}
    \max_{\theta_f: W^z_k \geq 0, \forall k} \min_{\theta_g} f_{\theta_f}(\nabla g_{\theta_g}(y)) - \langle y, \nabla g_{\theta_g}(y) \rangle - f_{\theta_f}(x) + \lambda R(\theta_g).
\end{align*}

\subsubsection{Projection Module}
For very high-dimensional inputs such as single-cell RNA seq data, we project the data into a lower-dimensional space.
The effect of a perturbation effect is then learned on the control particles encoded into a lower dimensional space. Subsequently, we decode the predicted target particles into the original data space.
We consider both, principal component (PCA) as well as autoencoder-based projections. When conducting experiments in PCA space, we consider the first 50 principal components, as they contain $> 99\,\%$ of the explained variance (see Fig.~\ref{fig:exp_action_norman_line}c).
The autoencoder architecture is inspired by \citep{lotfollahi2019scgen}, as it has been designed and tested for single-cell RNA seq data. The results reported in \S~\ref{sec:evaluation} are based on autoencoder projections and the evaluation metrics are computed on the decoded target particles.

\subsection{Hyperparameters} \label{app:hyperparams}

\looseness -1 To learn the optimal transport maps, we use PICNN architectures of 4 hidden layers of width 64.
The autoencoder parameterizing the projection module consists of an encoder and decoder with each 2 layers of 512 dimensional hidden layers. The size of the latent space is 50. The deep set consists of an encoder with 2 linear layers with 8 hidden units, followed by a \texttt{sum}-pooling operator and a 2 layer decoder with 8 hidden units, returning a set embedding of the same size as each individual input embedding, and passed through a final sigmoid activation function.
For all networks, we use the Adam optimizer \citep{kingma2014adam} with a learning rate of $0.0001$ ($\beta_1=0.5$, $\beta_2=0.9$) and $\lambda$=1.
If $\mathcal{T}$ is learned via the OT dual, $f$ and $g$ are learned via an alternate min-max optimization. $f$ is updated by fixing $g$ and maximizing \eqref{eq:makkuva_f_loss} with a single iteration. Then, for 10 iterations, i.e., \texttt{train\_freq\_f}$ = 10$, $f$ is fixed, and $g$ is optimized by minimizing \eqref{eq:makkuva_g_loss}. For the baselines, we followed the default configurations specified by the authors on the same datasets. We use a default batch size of 256, which is adapted for perturbations with fewer cells (due to a train / test split of $80\% \slash 20\%$).

\section{Reproducibility}
An implementation of \textsc{CondOT} is available at \href{https://github.com/bunnech/condot}{github.com/bunnech/condot}.

\end{document}